\documentclass[runningheads]{llncs}
\usepackage[T1]{fontenc}
\usepackage[pdftex]{graphicx}
\usepackage{color}
\usepackage{amsmath}

\usepackage{wrapfig}
\usepackage{algorithm}
\usepackage{algorithmicx} 
\usepackage[noend]{algpseudocode}
\usepackage[hyphens]{url}
\usepackage{colortbl}
\usepackage{subcaption}
\usepackage{makecell}
\usepackage{pifont}

\usepackage{xcolor}
\usepackage{booktabs}
\usepackage{array}
\usepackage{stackengine}
\usepackage[export]{adjustbox}
\usepackage[toc,page]{appendix}

\definecolor{Gray}{gray}{0.9}
\usepackage[pagebackref,breaklinks]{hyperref}
\usepackage[capitalize, nameinlink]{cleveref}
\usepackage{enumitem}

\usepackage[ruled,linesnumbered,vlined,algo2e]{algorithm2e} 
\usepackage{subcaption}
\usepackage{tabularx}
\usepackage{multirow}
\usepackage{multicol}

\newcolumntype{B}{>{\centering}p{0.06\textwidth}}
\newcolumntype{U}{>{\centering\arraybackslash}p{0.098\textwidth}}
\newcolumntype{W}{>{\centering}p{0.098\textwidth}}
\newcolumntype{C}{>{\centering}p{0.13\textwidth}}
\newcolumntype{T}{>{\centering}p{0.16\textwidth}}
\newcolumntype{I}{>{\centering}p{0.14\textwidth}}
\newcolumntype{E}{>{\centering\arraybackslash}p{0.14\textwidth}}

\newcolumntype{Z}{p{0.06\textwidth}}
\newcolumntype{A}{>{\raggedleft}p{0.08\textwidth}}
\newcolumntype{D}{>{\raggedleft\arraybackslash}p{0.085\textwidth}}

\usepackage{etoolbox}
\makeatletter
\patchcmd{\hyper@makecurrent}{%
    \ifx\Hy@param\Hy@chapterstring
        \let\Hy@param\Hy@chapapp
    \fi
}{%
    \iftoggle{inappendix}{
        \@checkappendixparam{chapter}%
        \@checkappendixparam{section}%
        \@checkappendixparam{subsection}%
        \@checkappendixparam{subsubsection}%
        \@checkappendixparam{paragraph}%
        \@checkappendixparam{subparagraph}%
    }{}%
}{}{\errmessage{failed to patch}}

\newcommand*{\@checkappendixparam}[1]{%
    \def\@checkappendixparamtmp{#1}%
    \ifx\Hy@param\@checkappendixparamtmp
        \let\Hy@param\Hy@appendixstring
    \fi
}
\makeatletter

\newtoggle{inappendix}
\togglefalse{inappendix}

\apptocmd{\appendix}{\toggletrue{inappendix}}{}{\errmessage{failed to patch}}
\apptocmd{\subappendices}{\toggletrue{inappendix}}{}{\errmessage{failed to patch}}

\begin{document}

\title{VFLIP: A Backdoor Defense for Vertical Federated Learning via Identification and Purification}
%
%

\author{Yungi Cho\inst{1}\orcidID{0000-0003-1297-8586} \and
Woorim Han\inst{1}\orcidID{0000-0003-0895-0986} \and 
Miseon Yu\inst{1}\orcidID{0000-0002-6076-1376} \and
Younghan Lee\inst{1}\orcidID{0000-0001-8414-966X} \and
Ho Bae\inst{2}\protect\footnote[1]{Correspondence should be addressed to H. Bae and Y. Paek.}\orcidID{0000-0002-5238-3547} \and
Yunheung Paek\inst{1}\protect\footnotemark[1]\orcidID{0000-0002-6412-2926}
}
\authorrunning{This paper has been accepted by ESORICS 24}
\titlerunning{VFLIP: A Backdoor Defense for Vertical Federated Learning}
%
\institute{Dept. of ECE and ISRC, Seoul National University, Seoul 08826, Republic of Korea\\
\email{\{q1w1ert1,rimwoo98,altjs543,201younghanlee,ypaek\}@snu.ac.kr}\\
\url{http://sor.snu.ac.kr}\\
\and
Dept. of Cyber Security, Ewha Womans University, Seoul 03760, Republic of Korea\\
\email{hobae@ewha.ac.kr}\\
\url{https://www.spai.co.kr/}\\
}
%

%
\maketitle              
\begin{abstract}
Vertical Federated Learning (VFL) focuses on handling vertically partitioned data over FL participants. Recent studies have discovered a significant vulnerability in VFL to \textit{backdoor attacks} which specifically target the distinct characteristics of VFL. Therefore, these attacks may neutralize existing defense mechanisms designed primarily for Horizontal Federated Learning (HFL) and deep neural networks. In this paper, we present the first backdoor defense, called \textit{VFLIP}, specialized for VFL. VFLIP employs the \textit{identification} and \textit{purification} techniques that operate at the inference stage, consequently improving the robustness against backdoor attacks to a great extent. VFLIP first identifies backdoor-triggered embeddings by adopting a participant-wise anomaly detection approach. Subsequently, VFLIP conducts purification which removes the embeddings identified as malicious and reconstructs all the embeddings based on the remaining embeddings. We conduct extensive experiments on CIFAR10, CINIC10, Imagenette, NUS-WIDE, and Bank-Marketing to demonstrate that VFLIP can effectively mitigate backdoor attacks in VFL. \url{https://github.com/blingcho/VFLIP-esorics24}

\end{abstract}

\section{Introduction}

\begin{figure*}

    \centering
    \includegraphics[width=0.60\textwidth]{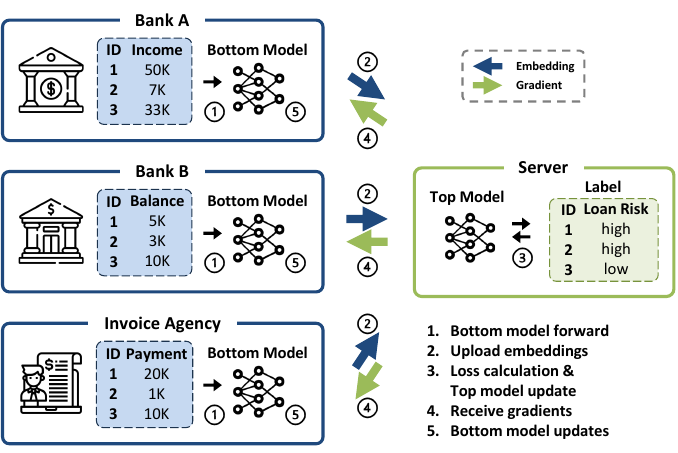}
    \caption{An illustration of VFL with the split neural network}
    \label{fig:vfl_overview}
    \vspace{-20pt}
\end{figure*}
Federated learning (FL) is a privacy-preserving machine learning framework that enables multiple participants to collaboratively train a model without directly sharing their private data. Instead, participants exchange local computations, such as model weights, gradients, and embeddings. FL can be categorized into two types based on the distribution of data among participants: Horizontal Federated Learning (HFL) and Vertical Federated Learning (VFL)~\cite{liu2023verticalconcepts}. In HFL, each participant has distinct sets of samples that share the same features. However, in VFL, participants handle the same samples, but each possesses a unique subset of features for these samples. In sectors where data privacy is of utmost importance, such as finance and healthcare, it is common for subsets of features to be distributed across multiple organizations~\cite{jin2021cafe}. In such situations, VFL offers a compelling solution by employing a split neural network architecture~\cite{fu2022label,bai2023villain,xuan2023badvfl,lai2023vfedad}. As illustrated in~\autoref{fig:vfl_overview}, each participant operates a~\textit{bottom model} tailored to its unique subset of features. The subsets are disjoint across participants yet pertain to the same set of samples (e.g., IDs 1,2,3). Instead of sharing the raw features with sensitive information, participants compute and share~\textit{embeddings} derived from the bottom model. The central server hosts a~\textit{top model}, which uses the embeddings to infer the labels of the samples.


FL confronts a range of security threats arising from the involvement of unreliable or malicious participants who deviate from the majority's intention ~\cite{wei2022vertical,liu2022copur,shejwalkar2021manipulating,xuan2023badvfl}. During the training stage, some participants may send malicious local computations to the server to manipulate the model's behavior. Relatively much work has been done to reduce the impact of such malicious computations in HFL~\cite{blanchard2017machine,lyu2020threats,fang2020local}. However, little has been conducted to defend against malicious participants in VFL. One of the most well-known security threats posed by the malicious participants in VFL is the \textit{backdoor attack}~\cite{bai2023villain,xuan2023badvfl}, where an attacker subtly manipulates the training data by planting a backdoor trigger during the model's training stage. The trigger is carefully designed to alter the predictions to a target label of the attacker's choice during inference. Sadly, we have discovered that the unique characteristics inherent in VFL make it difficult to apply the HFL defense mechanisms, which deal with computations derived from samples that share the same features~\cite{nguyen2022flame,blanchard2017machine}. Moreover, these challenges cannot be effectively mitigated by existing backdoor defense mechanisms designed for DNNs~\cite{bai2023villain,liu2018fine,wu2021adversarial,li2021anti,li2021backdoor}. Consequently, there is an urgent need for a defense mechanism tailored to countering backdoor attacks specialized for VFL.

The first challenge~\textbf{(C1)} in designing the defense mechanism for VFL is that the server aggregates~\textit{embeddings} from the participants to predict the label of a given sample in the inference stage~\cite{liu2023verticalconcepts}. This unique aspect of the split neural network architecture inherent in VFL introduces a new attack surface that can be exploited by attackers. By exploiting the newly exposed attack surface, a recent work~\cite{bai2023villain} proposed a novel backdoor attack with~\textit{embedding-level} backdoor triggers. They demonstrated that such attacks can be both effective and stealthy, creating a robust backdoor trigger that manipulates the model's prediction to a targeted label. 
The second challenge~\textbf{(C2)} stems from the unique nature of vertically partitioned feature configuration in VFL. Such configuration significantly complicates the process of detecting attackers among participants. In HFL, each local update (e.g., model weights or gradients) is computed from data with the same feature space. This allows direct comparison of local updates to identify attackers who deviate from the majority of participants~\cite{nguyen2022flame,blanchard2017machine,fang2020local}. In contrast, VFL presents a different setting where each participant only possesses a subset of features. Each embedding computed from a different feature subset is difficult to compare. Thus, it is necessary to design a new approach to identify malicious embeddings, which differ from most embeddings. 

\begin{figure}[t!]
    \centering
    \includegraphics[width=0.65\textwidth]{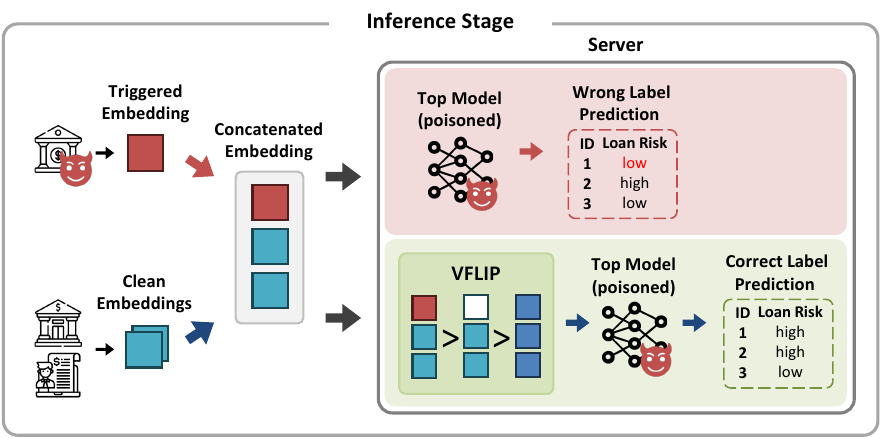}
    \caption{A brief summary of VFLIP. VFLIP identifies the backdoor-triggered embeddings and purifies all the embeddings through removal and reconstruction.}
    \label{fig:overview}
    \vspace{-15pt}
\end{figure}

    

To address these challenges, we propose a backdoor defense for VFL via Identification and Purification (VFLIP) at the inference stage, which employs a Masked Auto-Encoder (MAE) as a key component. VFLIP consists of two phases: identification and purification. In the \textbf{identification} phase, we introduce a participant-wise anomaly detection method with the majority voting to resolve the above challenges~\textbf{(C1-2)}. This approach is inspired by the attack strategy of the previous study~\cite{bai2023villain,xuan2023badvfl}. In these studies, as the participants lack the ability to manipulate the labels on the server, a two-fold approach is employed. During training, the trigger is injected into samples with the target label establishing a connection between the trigger and the target label. In contrast, for inducing misclassification during inference, the trigger is introduced into samples associated with non-target labels. Given the inter-participant correlations within the sample, as highlighted in previous studies~\cite{lai2023vfedad,liu2022copur,gharibshah2022local}, the embedding correlation among participants during training inevitably diverges from that observed during the inference stage. Such discrepancy allows the training of MAE to incorporate an anomaly detection approach during the inference stage, facilitating the identification of abnormal relations caused by malicious participants. Thus, VFLIP identifies the embeddings that show abnormal relationships from most other embeddings. Subsequently, we conduct the~\textbf{purification} phase using the MAE to minimize the impact of the backdoor-triggered embeddings. This process removes the identified backdoor-triggered embeddings and reconstructs all of the embeddings based on the remaining embeddings. 

Capitalizing on the denoising capability of the MAE~\cite{vincent2008extracting}, VFLIP adeptly reconstructs all the embeddings, minimizing the influence of malicious embeddings with stealthy triggers.~\autoref{fig:overview} presents a brief summary of VFLIP, showing that even when an attacker attempts to manipulate the model prediction by providing a backdoor-triggered embedding, the VFL model with VFLIP can predict the correct label. Our main contributions are as follows.

\begin{enumerate}
\item We propose VFLIP, a simple yet powerful method for defending against backdoor attacks in VFL, which conducts participant-wise anomaly detection with majority voting. To the best of our knowledge, this is the first study to defend against the backdoor attacks specialized for VFL with split architecture.

\item We conduct extensive experiments on CIFAR10, CINIC10, Imagenette, NUS-WIDE, and Bank-Marketing. This demonstrates that VFLIP effectively defends against the state-of-the-art attack methods by reducing the attack success rate from 84.4\% to 7.57\% on average.

\item We design an adaptive attack strategy for compromising the VFLIP's MAE. Through this, we demonstrate that it is hard for the attackers to compromise the MAE without significantly decreasing their attack performance.
\end{enumerate}

\section{Preliminaries}

\subsection{Vertical Federated Learning}
The fundamental concept of VFL with a split neural network is dividing the model into two parts: bottom models that take local data as inputs and produce embeddings, and a top model that makes a final decision based on the embeddings from the participants~\cite{fu2022label,bai2023villain}. Each participant has a bottom model and a subset of the joint features, while the server holds the top model and the labels.  
Following the previous VFL setup~\cite{fu2022label,bai2023villain}, we suppose that there are $N$ participants and a server, with the collaborative goal of training a model and subsequently performing inference on a sample using the trained model. VFL model training is conducted on a dataset $\mathcal{D}=\{(x^{k},y^{k})\}^{K}_{k=1}$ where $x$ represents a joint data sample, $y$ is the corresponding label, $K$ is the total number of data samples, and $k$ is the index for each data sample. In the feature-partitioned environment, a joint data sample can be expressed as $x = [x_1,\cdots,x_n]$. The $i$-th participant holds a vertically partitioned local dataset, denoted as $\mathcal{D}_i=\{x_i^{k}\}^{K}_{k=1}$. The $i$-th participant's bottom model $B_i$ maps local data $x_{i}$ to the embedding $h_{i}$. For simplicity, the parameters of the bottom models are denoted as $\theta_{B_1,...,B_N}$. The server owns the top model $T(h_1,\cdots,h_{N})$ parameterized as $\theta_{T}$. We denote the loss function of VFL as $\mathcal{L}$. The objective function can be formulated as follows:
\begin{equation} \label{eqn:one}
    \underset{\theta_{\operatorname{B}_1,...,\operatorname{B}_N}, \theta_{\operatorname{T}} }{\operatorname{argmin}}\; \sum_{k=1}^{K} \mathcal{L}(T([B_1(x_1^k),\cdots,B_{N}(x_{N}^k)]),y^k) 
\end{equation}

The training stage for VFL consists of five main steps: 
1) \textbf{Batch index selection}: The server selects indices, denoted as $idx$, from $\mathcal{D}$ and shares it with the VFL participants; 2) \textbf{Bottom model forward pass}: Each participant computes their embeddings $h_i^{idx}$ with $B_i(x_i^{idx})$ and sends it to the server; 3) \textbf{Top model forward pass}: The server concatenates all embeddings from participants corresponding to $idx$ and computes model prediction through $T(h_1^{idx},\cdots,h_{N}^{idx})$; 4) \textbf{Top model backward propagation}: The server calculates the loss with labels. Using the loss, the server computes the gradients of the top model and updates it. Afterward, the server sends back the gradients associated with the participants' embeddings; 5) \textbf{Bottom model backward propagation}: Each participant performs backward propagation using the gradients received from the server and updates their bottom models. This process is repeated during the training stage. For the inference stage, steps 1 to 3 are executed, and in step 1, $idx$ is chosen from the test set, not the training set $\mathcal{D}$. We note that VFLIP can be applied in scenarios where the server possesses features and participates in training with their own bottom model. This paper focuses on scenarios where the server cannot access the features of the training data.

\subsection{Backdoor Attacks in VFL}
The objective of a backdoor attack is to manipulate the model so that it correctly predicts clean samples but misclassifies backdoor-triggered samples as the target label~\cite{gu2017badnets,xuan2023badvfl,bai2023villain}. Depending on whether the attacker can manipulate the labels of the training set, backdoor attacks can be divided into clean-label attacks and dirty-label attacks. The clean-label backdoor attack injects a trigger only into the samples of the target label~\cite{shafahi2018poison,huang2020metapoison}. 
On the other hand, the dirty-label attack injects a trigger into the samples of non-target labels and manipulates the labels to the target label~\cite{gu2017badnets}. Since the labels on the server cannot be manipulated by the attackers in VFL, it is only susceptible to clean-label backdoor attacks~\cite{bai2023villain,xuan2023badvfl}.

\begin{figure*}[t!]
    \centering
    \includegraphics[width=\textwidth]{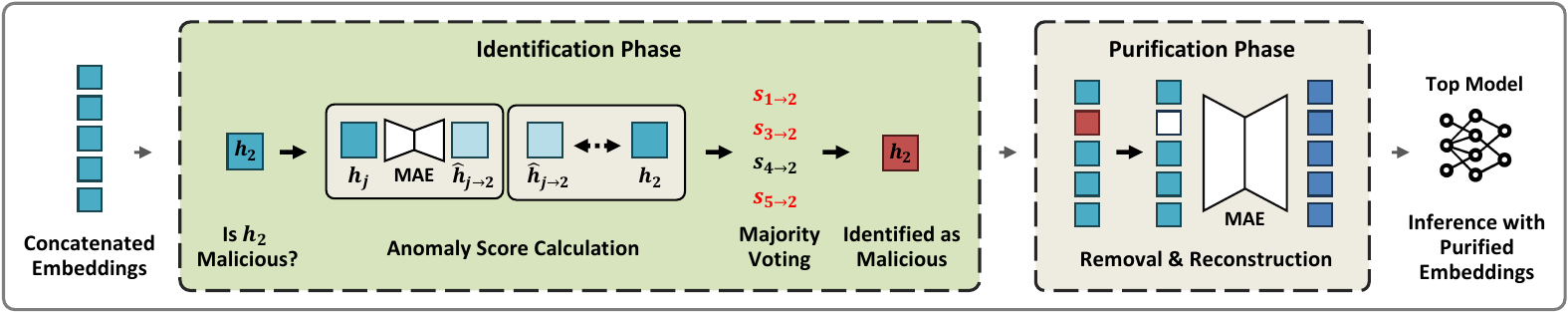}
    \caption{An overview of VFLIP. VFLIP calculates the anomaly scores for each embedding with the MAE. The voting mechanism is conducted based on the anomaly scores to determine whether an embedding is malicious. Embeddings identified as malicious are removed, and then all the embedding is reconstructed through MAE.}
    \label{fig:method}
\end{figure*}


Recent studies have proposed backdoor attacks tailored for VFL. Xuan \textit{et al.}~\cite{xuan2023badvfl} propose BadVFL, a data-level backdoor attack where the attacker plants backdoor triggers in their local data to manipulate the top model. To strengthen the connection between the trigger and the target label, the attacker replaces the local data of the target label with that of non-target labels before injecting a trigger. Bai \textit{et al.}~\cite{bai2023villain} introduce VILLAIN, which proposes an additive backdoor trigger on the embedding level, aiming for a stealthy backdoor attack. By adjusting the magnitude of this trigger, the attacker can control the trade-off between stealthiness and attack power. The details of the attack methods are provided in ~\autoref{Appendix:A}.

\subsection{Threat Model}
We follow the threat model of previous studies~\cite{bai2023villain,xuan2023badvfl}, but introduce a few modifications to consider strong attackers. There are one or more attackers among the VFL participants, but the number of attackers is limited to less than half of the total number of participants based on previous FL studies~\cite{liu2022copur,blanchard2017machine}. The attacker can modify their local data or embedding. This potentially allows the attacker to poison a substantial portion of the training set. However, they cannot modify any operation on the server or the benign participants. The attacker either has or does not have the label knowledge for their training set. The attacker without the label knowledge owns a small amount of labeled auxiliary data to infer labels about their training set~\cite{fu2022label,bai2023villain,xuan2023badvfl}. We note that even if the attacker knows the labels for the training set, they cannot change the labels in the server. The defender is the server that owns the top model and the labels. The server cannot access participants' operations.

\section{Method}
This section introduces VFLIP, a novel backdoor defense method for VFL. The primary objective of VFLIP is to diminish the influence of attackers who upload a backdoor-triggered embedding during the inference stage. VFLIP achieves this by identifying the backdoor-triggered embedding and purifying the concatenated embedding before feeding them to the top model. Leveraging the capability of the MAE for both tasks, VFLIP exhibits a mechanism illustrated in~\autoref{fig:method}.

\textbf{Identification} 
Using the discrepancy between the embedding correlation of training and inference stage made by conducting backdoor attacks in VFL, VFLIP detects the abnormal relationships between the embeddings of participants for each sample that are not observed during training. 
Here, the anomaly detection approach is applied by training the MAE with embeddings from the training stage. Subsequently, during the inference stage, the trained MAE is used to identify the backdoor-triggered embeddings that exhibit abnormal relationships with the majority of other embeddings.

\textbf{Purification}. 
The purification phase aims to mitigate the impact of the backdoor-triggered embedding by leveraging the identification results. To achieve this goal, VFLIP removes the embeddings identified as backdoor-triggered. Subsequently, VFLIP feeds the remaining embeddings to the MAE and reconstructs all the embeddings. Leveraging the denoising capability of the MAE~\cite{vincent2008extracting}, this process is particularly effective in alleviating the influence of malicious participants employing small-magnitude triggers.

The following subsections describe the MAE training and provide details about the identification and purification phase that utilizes the MAE.

\begin{algorithm2e}[t]
  \caption{VFLIP MAE Training} \label{alg:train_MAE}
  \DontPrintSemicolon
  \setcounter{AlgoLine}{0}
  
  \SetKwInOut{Input}{Input}\SetKwInOut{Output}{Output}
  \Input{~Masked Auto-Encoder MAE, Training set $\mathcal{H}^{\operatorname{train}}$, Training epoch $E_{\operatorname{mae}}$, Learning rates $\beta_1$ (for "$N$-$1$ to $1$") and $\beta_2$ (for "$1$ to $1$"), Masks $m$ filled with 1 for each local embedding, Number of participants $N$}
  \Output{~Trained MAE, Threshold $t$ for the anomaly score}

  \vspace{5pt}

  \textbf{Initialize} weights and bias for MAE

  \For{$\text{each}$ train epoch $e = 1,2,...,E_{\operatorname{mae}}$} {
    
    \For{each minibatch $B$ from $\mathcal{H}^{\operatorname{train}}$ } {
    
        \tcc{"$N$-$1$ to $1$" strategy}
        
        $\mathcal{L_{\operatorname{N-1}}} \gets 0$

        \For{each embedding $h$ from $B$ } {
        
            $h_i \gets$ randomly draw one local embedding from $h$

            $m_i \gets$ Mask filled with 1 for $h_i$

            $\Tilde{m}_{i} \gets 1 - m_i$
            
            $\hat{h} \gets \operatorname{MAE}(\Tilde{m}_{i}\odot h)$
            
            $\mathcal{L_{\operatorname{N-1}}} \gets \mathcal{L_{\operatorname{N-1}}} + ||m_{i}\odot(h - \hat{h}) ||_2 $

        }
    
        Update MAE to minimize $\mathcal{L_{\operatorname{N-1}}}$ with learning rate $\beta_1$
        
        \vspace{5pt}
        
        \tcc{"$1$ to $1$" strategy}
        
        $\mathcal{L_{\operatorname{1}}} \gets 0$

        \For{each embedding $h$ from $B$ } {
        
            $h_i, h_j \gets$ randomly draw two local embeddings from $h$

            $m_i, m_j \gets$ Mask filled with 1 for $h_i$, $h_j$
            
            $\hat{h} \gets \operatorname{MAE}(m_{j}\odot h)$
            
            $\mathcal{L_{\operatorname{1}}} \gets \mathcal{L_{\operatorname{1}}} + ||m_{i}\odot(h - \hat{h}) ||_2 $

        }
        Update MAE to minimize $\mathcal{L_{\operatorname{1}}}$ with learning rate $\beta_2$
    }
    }

    Compute the thresholds $t = [t_1, ..., t_N]$ for each participant over the $\mathcal{H}^{\operatorname{train}}$

  \Return Trained MAE, Threshold $t$
  
\end{algorithm2e}


\subsection{MAE Training}\label{subsec:train_MAE}
Initially, VFLIP trains the MAE, parameterized as $\theta_{\operatorname{MAE}}$, with the poisoned training set after the training stage of VFL. The training set for the MAE is embeddings gathered from the last epoch of the VFL training stage, denoted as $\mathcal{H}^{\operatorname{train}}$. The input of the MAE is the concatenated embedding, denoted as $h = [h_1, \cdots, h_{N}]$. The MAE outputs the reconstructed concatenated embedding, which is represented as $\hat{h} = [\hat{h}_{1}, \cdots, \hat{h}_{N}]$. The architecture of MAE is composed of an encoder and a decoder. Both of them use a fully connected network.

\subsubsection{Training strategies} VFLIP employs two MAE training strategies: "$N$-$1$ to $1$" and "$1$ to $1$". The "$N$-$1$ to $1$" strategy randomly chooses one embedding $h_i$ to be restored (Line 4-6 in \autoref{alg:train_MAE}). The selected $h_i$ is masked from $h$, and MAE reconstructs $\hat{h}_i$ using the masked $h$ (Line 7-9 in \autoref{alg:train_MAE}). Next, the "$1$ to $1$" strategy randomly selects two embeddings, denoted as $h_i$ and $h_j$ (Line 14 in \autoref{alg:train_MAE}). $h_i$ is the target to restore, and $h_j$ is used as the input for restoration. Here, all the embeddings except $h_j$ are masked from $h$, and MAE reconstructs $\hat{h}_i$ using the masked $h$ (Line 15-17 in \autoref{alg:train_MAE}). 

Following the above strategies, the loss is calculated only for the selected embedding $h_i$ (Line 10, 17 in \autoref{alg:train_MAE}). MAE is trained by alternately optimizing the loss for the two strategies (Line 11, 18 in \autoref{alg:train_MAE}). The objective functions for each strategy are as follows:

\begin{equation} \label{eqn:mask1}
    \underset{\theta_{\operatorname{MAE}}}{\operatorname{argmin}}\;   
     ||m_{i}\odot(h - \operatorname{MAE}(\Tilde{m}_{i}\odot h)) ||_2 
\end{equation}
\begin{equation} \label{eqn:mask2}
    \underset{\theta_{\operatorname{MAE}}}{\operatorname{argmin}}\;   
     ||m_{i}\odot(h - \operatorname{MAE}(m_{j}\odot h))) ||_2
\end{equation}

Here, $m_i$ represents a masking value where only the part corresponding to $h_i$ is filled with 1, while the rests are filled with 0. $\Tilde{m_i}$ represents the opposite of $m_i$, where 0s and 1s are reversed. ~\autoref{eqn:mask1} and ~\autoref{eqn:mask2} are for "$N$-$1$ to $1$" and  "$1$ to $1$" strategy, respectively. The ablation study for these training strategies is provided in ~\autoref{sec:ablation}.

\subsubsection{Standardization and drop-out}
To enhance the MAE performance, VFLIP employs standardization and drop-out. Standardization is a data preprocessing technique where, instead of directly using $h$ as input, MAE uses standardized $h$. VFLIP's standardization is based on the mean and standard deviation of $\mathcal{H}^{\operatorname{train}}$. Additionally, drop-out is used for data augmentation. During training, randomly generated masks partially remove $h$ to prevent MAE from overfitting to specific dimensions.

\subsection{VFLIP mechanism}
\begin{algorithm2e}[t]
  \caption{VFLIP mechanism} \label{alg:byzan}
  \DontPrintSemicolon
  \setcounter{AlgoLine}{0}
  
  \SetKwInOut{Input}{Input}\SetKwInOut{Output}{Output}
  \Input{~Concatenated embedding $h$, Trained Masked Auto-Encoder MAE, Masks $m$ filled with 1 for each embedding, Number of participants $N$, Thresholds $\{t_i\}_{i=1}^{N}$ for the anomaly score, VFL top model $T$}
  \Output{~VFL model prediction $P$}
  
 \vspace{5pt}

  \textbf{Initialize} $votes[N]$ with $0$

  \tcc{$votes$ is an array for counting votes}
  
  \textbf{Initialize} $m_{\operatorname{mal}}$ with $[0,...,0]$

 \tcc{$m_{\operatorname{mal}}$ is a mask for malicious participants' part}
 
  \vspace{5pt}

   \tcc{Identifying the backdoor-triggered embedding}

  \For{each local embedding $h_i$ = $h_1,...,h_N$} {
    \For{each local embedding $h_j$ = $h_1,...,h_N$} {
        \If{$h_i$ is not $h_j$}{
            $s_{j\rightarrow i} \gets ||m_{i}\odot(h - \operatorname{MAE}(m_{j}\odot h) ||_2$
            
            \If{$s_{j\rightarrow i} > t_i$}{
                $votes[i] \gets votes[i] + 1$
            }
        }
     }
  
     \If{$votes[i] > \frac{N}{2}$}{
                $m_{\operatorname{mal}} \gets m_{\operatorname{mal}} + m_i$
            }
    }

\vspace{5pt}
  \tcc{Purifying the concatenated embedding}

 $h_{\operatorname{removed}} \gets (1-m_{\operatorname{mal}}) \odot h $
 
 $\hat{h}_{\operatorname{purified}} \gets \operatorname{MAE}(h_{\operatorname{removed}})$

 $P \gets T(\hat{h}_{\operatorname{purified}}) $
\vspace{5pt}
  
  \Return VFL model prediction  $P$
\end{algorithm2e}
\subsubsection{Identification}
In the identification phase, VFLIP conducts the participant-wise anomaly detection with majority voting. In this process, anomaly scores for one participant's embedding are calculated from other participants' embeddings. Then, based on these anomaly scores, the participant's embedding is determined whether it is malicious or benign through majority voting. This process is conducted for each participant's embedding.

\textbf{Anomaly score calculation}. For an embedding of one participant, VFLIP calculates N-1 anomaly scores based on the embeddings of other N-1 participants (Line 6 in \autoref{alg:byzan}). To be specific, for $h_i$, VFLIP masks all the embeddings except for $h_j$ in the concatenated embedding and reconstructs $\hat{h}$ using the MAE. The part of $\hat{h}$ corresponding to $h_i$, based on $h_j$, is denoted as $\hat{h}_{j\rightarrow i}$. VFLIP defines the anomaly score $s_{j\rightarrow i}$ as follows:

\begin{equation} 
    \hat{h}_{j\rightarrow i} = m_{i}\odot\operatorname{MAE}(m_{j}\odot h) 
\end{equation}
\begin{equation} 
    s_{j\rightarrow i} = ||\hat{h}_{j\rightarrow i} - h_i ||_2
\end{equation}

To understand this anomaly score, it is essential to delve into the characteristics of VFL backdoor attacks~\cite{bai2023villain,xuan2023badvfl}. The attacker injects a backdoor trigger only into the target label samples during the training stage. Therefore, at the inference stage, if the attacker inserts the backdoor trigger into non-target label samples to manipulate the VFL model predictions, it results in a relatively high anomaly score for the attacker's embedding because the MAE incorrectly generates the attacker's embedding as the MAE did not learn about the relationships between the backdoor-triggered embedding and the embeddings of the non-target labels. 

\textbf{Majority voting}. If the anomaly score $s_{j \rightarrow i}$ exceeds the threshold $t_i$, $h_i$ gets a vote (Line 7-8 in \autoref{alg:byzan}). The threshold $t_i$ is determined by the $\mu_{i} + \rho\cdot\sigma_i$, where $\mu_i$ and $\sigma_i$ represent the mean and standard deviation of all anomaly scores from $\mathcal{H}^{\operatorname{train}}$ for $i$-th participant. Since VFLIP aims to detect abnormal cases that are not observed in the VFL training stage, it is reasonable to set the threshold based on the distribution of anomaly scores obtained from $\mathcal{H}^{\operatorname{train}}$. $\rho$ is the hyper-parameter for controlling the threshold. If the number of votes $h_i$ received is greater than half of the number of total participants ($\frac{N}{2}$), it is considered as a backdoor-triggered embedding (Line 9-10 in \autoref{alg:byzan}). 

\subsubsection{Purification}
VFLIP removes the malicious embeddings from the concatenated embedding (Line 11 in \autoref{alg:byzan}) and feeds them into MAE to obtain the purified $\hat{h}$ (Line 12 in \autoref{alg:byzan}). Subsequently, the top model uses $\hat{h}$ as input to obtain the final prediction  (Line 13 in \autoref{alg:byzan}). 
\section{Experiments}
\subsection{Experiments Setup}
\textbf{Dataset descriptions}. Following the previous VFL studies~\cite{xuan2023badvfl,bai2023villain,liu2022copur}, we evaluate the effectiveness of VFLIP using five datasets: three image datasets (i.e., CIFAR10~\cite{Krizhevsky09learningmultiple}, CINIC10~\cite{darlow2018cinic10}, Imagenette~\cite{howard2020fastai}), one image-text combined dataset (i.e., NUS-WIDE~\cite{chua2009nus}), and one financial dataset (i.e., Bank Marketing (BM)~\cite{MORO201422}). CIFAR10 and CINIC10 have 10 classes consisting of 32x32 pixel images. Imagenette is a 10-class subset of the Imagenet dataset. Each image is resized by 224x224 pixels. NUS-WIDE has 81 classes with 634 image features and 1000 text features. We select the five classes following the previous study~\cite{liu2022copur}. BM has two classes with 40 features.

\textbf{Default training setup}. We validate VFLIP under a four-participant scenario with a single attacker and an eight-participant scenario with three attackers. Based on previous VFL studies~\cite{fu2022label,bai2023villain}, the data features are vertically split among the participants. The optimization method is Stochastic Gradient Descent (SGD). The bottom model architecture is VGG19~\cite{simonyan2014very} for the image datasets and a 4-layer fully connected network (FCN) for NUS-WIDE and BM. The top model uses a 3-layer FCN. The VFL training epoch is set to 50 for CIFAR10, CINIC10, Imagenette, and NUS-WIDE, and is set to 40 for BM.

\textbf{Attacks}. We evaluate VFLIP on two SOTA attacks: BadVFL~\cite{xuan2023badvfl} and VILLAIN~\cite{bai2023villain}. The attacker without label knowledge conducts the label inference attack proposed by the previous study~\cite{bai2023villain}. The attacker injects a backdoor trigger after $E_{bkd}$ epoch. For the attacker without label knowledge, the poisoning budget is set to 10\% and $E_{bkd}$ is set to 20. For the attacker with label knowledge, the poisoning budget is set to 50\% and $E_{bkd}$ is set to 5. The attacker's target label is set to 0. The details for attack settings are provided in ~\autoref{app:hyper_attack}. 

\begin{table}[t]
    \caption{Evaluation for a single attacker on five datasets. No DEF (no defense): Result without any defense mechanism.}
    \label{tbl:table_single}
    \begin{center}
         \normalsize
\def\arraystretch{1.15}
\resizebox{\textwidth}{!}{
\begin{tabular}{|T|I|I||WWWWWW||WWWWWU|}
\hline
        \multirow{3}{*}[-0.7em]{Dataset} & \multirow{3}{*}[-0.6em]{\makecell{Label\\ Knowledge}} & \multirow{3}{*}[-0.7em]{Attack} & \multicolumn{12}{c|}{Defense} \\

        \cline{4-15}
        
        {} & {} & {} & \multicolumn{6}{c||}{Accuracy (\%) $\uparrow$  \text{(Higher is better)}} & \multicolumn{6}{c|}{Attack Success Rate (\%) $\downarrow$ \text{(Lower is better)}} 
        \\
        \cline{4-15}
        
        {} & {} & {} & NO DEF & DP-SGD & \multirow{1}{*}[-0.7em]{MP}  & \multirow{1}{*}[-0.7em]{ANP} & \multirow{1}{*}[-0.7em]{BDT}  & \multirow{1}{*}[-0.7em]{VFLIP} & NO DEF & DP-SGD & \multirow{1}{*}[-0.7em]{MP}  & \multirow{1}{*}[-0.7em]{ANP} & \multirow{1}{*}[-0.7em]{BDT}  & \multirow{1}{*}[-0.7em]{VFLIP} \\
        \hline
        \hline

        \multirow{4}{*}[-0.3em]{CIFAR10}  & \multirow{2}{*}{w/o}  & BadVFL 
        &77.34	&\textbf{78.04}	&75.35	&75.53	&73.28	&75.14	
        &30.58	&30.81	&25.55	&28.14	&29.20	&\textbf{13.95}
        \\

        {} & {}  & VILLAIN 
        &\textbf{76.84}	&75.04	&75.34	&73.40	&73.15	&75.22	
        &86.96	&20.40	&79.17	&64.83	&83.00	&\textbf{3.19}
        \\
        \cline{2-15}
        
        {} & \multirow{2}{*}{with}  & BadVFL 
        &76.85	&\textbf{77.30}	&74.83	&75.41	&72.46	&75.46	
        &97.50	&99.99	&87.13	&96.49	&95.51	&\textbf{5.52}
        \\

        {} & {}  & VILLAIN 
        &75.46	&\textbf{75.52}	&74.94	&74.11	&74.33	&73.86	
        &99.84	&64.13	&99.76	&99.80	&99.77	&\textbf{2.79}
        \\

        \hline
        \hline
        \multirow{4}{*}[-0.3em]{CINIC10}  & \multirow{2}{*}{w/o}  & BadVFL 
        &\textbf{64.60}	&64.43	&64.37	&64.34	&61.94	&62.45	
        &21.13	&28.27	&18.05	&17.89	&19.52	&\textbf{13.90}
        \\

        {} & {}  & VILLAIN 
        &\textbf{63.66}	&63.14	&63.30	&63.33	&60.88	&62.27	
        &75.65	&11.53	&64.30	&65.03	&70.36	&\textbf{3.74}
        \\
        \cline{2-15}
        
        {} & \multirow{2}{*}{with}  & BadVFL 
        &\textbf{64.37}	&62.46	&62.22	&63.88	&60.40	&63.21	
        &99.46	&80.31	&91.51	&96.82	&99.25	&\textbf{4.92}
        \\

        {} & {}  & VILLAIN 
        &\textbf{62.43}	&61.73	&61.93	&62.09	&60.91	&61.05	
        &99.97	&85.41	&99.91	&99.94	&99.97	&\textbf{2.34}
        \\

        \hline 
        \hline
        \multirow{4}{*}[-0.3em]{Imagenette} & \multirow{2}{*}{w/o}  & BadVFL 
        &\textbf{75.48}	&74.98	&72.54	&75.07	&71.22	&73.37	
        &61.02	&78.01	&54.37	&60.21	&58.46	&\textbf{14.41}
        \\

        {} & {}  & VILLAIN 
        &\textbf{74.11}	&73.83	&72.55	&73.12	&71.25	&71.60	
        &96.66	&23.80	&93.95	&96.50	&96.50	&\textbf{2.87}
        \\
        \cline{2-15}
        
        {} & \multirow{2}{*}{with}  & BadVFL 
        &74.02	&\textbf{74.97}	&70.84	&73.64	&69.43	&71.70	
        &92.21	&93.20	&73.36	&92.08	&90.62	&\textbf{3.79}
        \\

        {} & {}  & VILLAIN 
        &72.31	&\textbf{73.27}	&69.50	&71.37	&69.45	&69.32	
        &97.38	&89.29	&96.70	&97.33	&97.42	&\textbf{1.87}
        \\

        \hline 
        \hline
        \multirow{4}{*}[-0.3em]{NUS-WIDE} & \multirow{2}{*}{w/o}  & BadVFL 
        &\textbf{83.55}	&83.32	&82.99	&81.42	&81.66	&81.26	
        &72.65	&57.68	&68.66	&61.82	&73.12	&\textbf{9.28}
        \\

        {} & {}  & VILLAIN 
        &\textbf{83.74}	&83.16	&81.78	&80.75	&81.21	&81.39	
        &89.34	&23.73	&80.13	&72.33	&88.75	&\textbf{7.65}
        \\
        \cline{2-15}
        
        {} & \multirow{2}{*}{with}  & BadVFL 
        &\textbf{82.79}	&81.45	&80.72	&82.52	&81.76	&79.93	
        &99.99	&95.55	&95.11	&99.98	&99.97	&\textbf{12.45}
        \\

        {} & {}  & VILLAIN 
        &\textbf{82.44}	&80.50	&80.47	&79.26	&80.68	&80.51	
        &100.00	&96.75	&99.91	&99.22	&100.00	&\textbf{6.30}
        \\

        \hline 
        \hline
        \multirow{4}{*}[-0.3em]{BM} & \multirow{2}{*}{w/o}  & BadVFL 
        &\textbf{93.74}	&93.70&	88.06&	93.58	&93.44&	90.35&	26.90&	22.25&	15.38&	31.39&	27.20&	\textbf{10.45}
        \\

        {} & {}  & VILLAIN 
        &\textbf{94.39}	&92.77	&91.41&	94.28&	90.23	&91.13	&45.98	&52.60	&31.24&	45.03&	44.94&	\textbf{5.59}
        \\
        \cline{2-15}
        
        {} & \multirow{2}{*}{with}  & BadVFL 
        &93.50	&93.67	&90.19	&\textbf{93.91}	&87.92	&91.75	
        &92.14	&64.41	&78.52	&94.30	&88.67	&\textbf{12.54}
        \\

        {} & {}  & VILLAIN 
        &93.73	&92.51	&90.50	&\textbf{93.75}	&89.51	&90.94
        &99.98	&55.93	&99.86	&100.00	&99.94	&\textbf{9.43}
        \\

        \hline

\end{tabular}
}

    \end{center}
    \vspace{-25pt}
\end{table}

\begin{table*}[t]
    \caption{Evaluation for multiple attackers with label knowledge on five datasets.}
    \label{tbl:table_multi}
    \begin{center}
\renewcommand{\arraystretch}{0.9}
\resizebox{\textwidth}{!}{
\begin{tabular}{|T|I||WWWWWW||WWWWWU|}
     \hline
        \multirow{3}{*}[-0.8em]{Dataset} & 
        \multirow{3}{*}[-0.8em]{Attack} & \multicolumn{12}{c|}{Defense} \\

        \cline{3-14}  
        
        {} & {} & \multicolumn{6}{c||}{Accuracy (\%) $\uparrow$  \text{(Higher is better)}} & \multicolumn{6}{c|}{Attack Success Rate (\%) $\downarrow$ \text{(Lower is better)}}
        \\
        
        \cline{3-8} \cline{9-14} 
        
        {} & {} & NO DEF & DP-SGD & \multirow{1}{*}[-0.65em]{MP}  & \multirow{1}{*}[-0.65em]{ANP} & \multirow{1}{*}[-0.65em]{BDT}  & \multirow{1}{*}[-0.65em]{VFLIP} & NO DEF & DP-SGD & \multirow{1}{*}[-0.65em]{MP}  & \multirow{1}{*}[-0.65em]{ANP} & \multirow{1}{*}[-0.65em]{BDT}  & \multirow{1}{*}[-0.65em]{VFLIP} 
        \\
        \hline
        \hline

        \multirow{2}{*}{CIFAR10} & BadVFL 
        &\textbf{74.57}	&74.31	&70.29	&72.99	&72.98	&70.59	&99.37	&100.00	&96.24	&99.16	&99.06	&\textbf{8.12}
        \\

        {}  & VILLAIN 
        &\textbf{71.58}	&70.56	&70.61	&71.30	&69.36	&68.44	&100.00	&94.57	&98.86	&100.00	&100.00	&\textbf{5.68}
        \\

        \hline 
        \hline
        \multirow{2}{*}{CINIC10}  & BadVFL 
        &\textbf{61.95}	&60.45	&58.65	&61.67	&57.87	&58.66	&99.88	&99.98	&91.07	&99.64	&99.81	&\textbf{5.37}
        \\

        {} & VILLAIN 
        &58.52	&55.98	&56.46	&\textbf{58.55}	&56.14	&55.98	&100.00	&98.09	&99.97	&100.00	&100.00	&\textbf{2.24}
        \\

        \hline 
        \hline
        \multirow{2}{*}{Imagenette}   & BadVFL 
        &\textbf{71.77}	&71.20	&70.85	&70.54	&67.04	&67.54	&98.50	&98.85	&96.85	&98.15	&97.96	&\textbf{3.00}
        \\

        {} & VILLAIN 
        &\textbf{69.71}	&69.01	&67.90	&69.03	&67.52	&66.13	&99.84	&96.20	&98.97	&99.69	&99.70	&\textbf{3.06}
        \\

        \hline 
        \hline
        \multirow{2}{*}{NUS-WIDE} &  BadVFL 
        &\textbf{81.81}	&78.80	&79.11	&81.44	&80.04	&78.67	&99.83	&97.50	&98.69	&99.85	&99.82	&\textbf{14.12}
        \\

        {}  & VILLAIN 
        &\textbf{82.76}	&75.54	&79.47	&81.35	&80.37	&80.62	&99.97	&80.21	&99.63	&99.57	&99.95	&\textbf{7.90}
        \\

        \hline 
        \hline
        \multirow{2}{*}{BM} &  BadVFL 
        &93.51	&93.73	&90.43	&\textbf{93.95}	&89.10 &91.01	
        &99.98	&99.09	&99.79	&100.00	&99.92	&\textbf{17.65}
        \\

        {}  & VILLAIN 
        &93.64	&92.07	&93.63	&\textbf{93.83}	&92.41	&90.28	
        &100.00 &96.88	&100.00	&100.00	&100.00	&\textbf{17.12}
        \\
        \hline
        
\end{tabular}

}

    \end{center}
\end{table*}

\textbf{Defense baselines}. To the best of our knowledge, there are no backdoor defenses specialized for VFL with split architecture. Following the previous study~\cite{bai2023villain}, we apply the existing backdoor defenses designed for DNNs, such as Model Pruning (MP)~\cite{liu2018fine}, Adversarial Neural Pruning (ANP)~\cite{wu2021adversarial}, and Backdoor Defense via Transform (BDT)~\cite{li2021backdoor} to defend against backdoor attacks in VFL and compare them with VFLIP. For applying BDT to the VFL scenario, we add noise to the embeddings following the previous study~\cite{bai2023villain}. 
Moreover, as DP-SGD~\cite{abadi2016deep} is known for improving the robustness against backdoor attacks to some extent~\cite{gao2023backdoor}, we analyze the ability of DP-SGD to defend against VFL backdoor attacks.

\textbf{Defense settings}.  
To find the defense hyperparameter for MP, ANP, and BDT in VFL, we explore various security-utility trade-off hyperparameters in a wide range like the previous study~\cite{min2023towards}, and report the results with the lowest ASR while maintaining accuracy. For MP, we vary the pruning ratios from 10\% to 90\%.
For ANP, similar to MP, we vary the ANP ratio from \%0.1 to 2\%.
For BDT, we increase the noise level until the accuracy is no longer maintained. 

In the case of DP-SGD, we select the lowest $\epsilon$ values among those showing stable model convergence: 6.35 for CIFAR10 and CINIC10, 4.65 for Imagenette and NUS-WIDE, and 8.84 for BM. The clipping value is determined based on the median of gradients' magnitude when DP-SGD is not applied, as guided by the previous study~\cite{abadi2016deep}. 

In VFLIP, the architecture of MAE employs a 3-layer FCN for both the encoder and the decoder. We set the learning rate of the "$N$-$1$ to $1$" strategy to 0.01 and the "$1$ to $1$" strategy's learning rate to 0.1. For identification, $\rho$ is set to 2 for CIFAR10, CINIC10, NUS-WIDE, and BM, and 2.5 for Imagenette. Dropout is set to 10\%. The MAE training epoch is set to 20 for CIFAR10, CINIC10, NUS-WIDE, and BM, and it is set to 50 for Imagenette.

\textbf{Metrics}. We employ two metrics to assess the robustness of VFLIP: clean accuracy (ACC) and attack success rate (ASR). ACC is the probability of correctly predicting the true label in the absence of backdoor triggers. ASR represents the probability of the model misclassifying the labels of the backdoor-triggered samples as the attacker's target label.

\subsection{Main Results}
~\autoref{tbl:table_single} presents the results of a four-participant scenario with a single attacker. The results for label inference accuracy are provided in~\autoref{app:able_label}. 
Each experiment is repeated five times with different seeds, and the average result is reported. In most cases, other defense techniques fail to mitigate backdoor attacks in VFL. On the other hand, VFLIP achieves the average ASR decrease of 85.56\%, 80.50\%, 91.84\%, 89.97\%, and 81.19\%, and the average ACC drop of 2.21\%, 2.38\%, 3.37\%, 2.84\%, and 3.36\% in CIFAR10, CINIC10, Imagenette, NUS-WIDE, and BM, respectively. Even though there is a slight accuracy drop compared with no defense, VFLIP demonstrates that it can effectively mitigate backdoor attacks in VFL. 

\subsection{Multiple Attackers}
To demonstrate the defense capabilities of VFLIP under multiple attackers, we conduct experiments in an eight-participant scenario with three label-knowledgable attackers, as presented in~\autoref{tbl:table_multi}. Each experiment is repeated five times with different seeds, and the average result is reported. The attackers share the information they possess, their objective, and their attack strategy and carry out all malicious actions simultaneously. Thus, the malicious portion in the concatenated embeddings is increased, making the attack more powerful. While other defense techniques fail to mitigate the backdoor attacks, for ASR, VFLIP achieves 6.9\%, 3.8\%, 3.03\%, 11.01\%, and 17.39\% on average for CIFAR10, CINIC10, Imagenette, NUS-WIDE, and BM, respectively. This indicates that VFLIP has the capacity to defend against multiple attackers. The ACC drop of VFLIP is 4.86\%, 4.83\%, 5.51\%, 3.21\%, and 3.13\% on average for CIFAR10, CINIC10, Imagenette, NUS-WIDE, and BM, respectively. Although there is a slight increase in ACC drop compared to the four-participant scenario, we note that VFL typically involves collaboration with a small number of participants~\cite{yang2019federated,wei2022vertical}, so situations resulting in an accuracy drop greater than observed in these experiments are rare.

\begin{figure}[t!]
  \centering
  \begin{subfigure}[b]{0.2\columnwidth}
    \centering
    \includegraphics[width=\columnwidth]{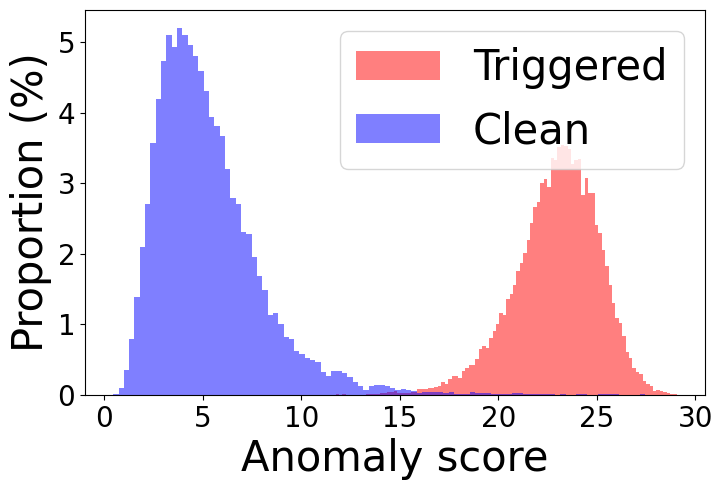} 
    \caption{CIFAR10}
    \label{fig:histogram_cifar_data}
  \end{subfigure}%
  \begin{subfigure}[b]{0.2\columnwidth}
    \centering
    \includegraphics[width=\columnwidth]{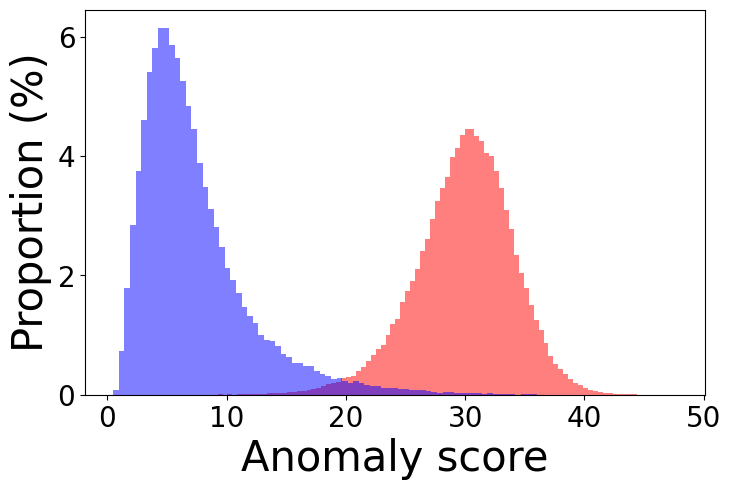}
    \caption{CINIC10}
    \label{fig:histogram_cinic_data}
  \end{subfigure}%
  \begin{subfigure}[b]{0.2\columnwidth}
    \centering
    \includegraphics[width=\columnwidth]{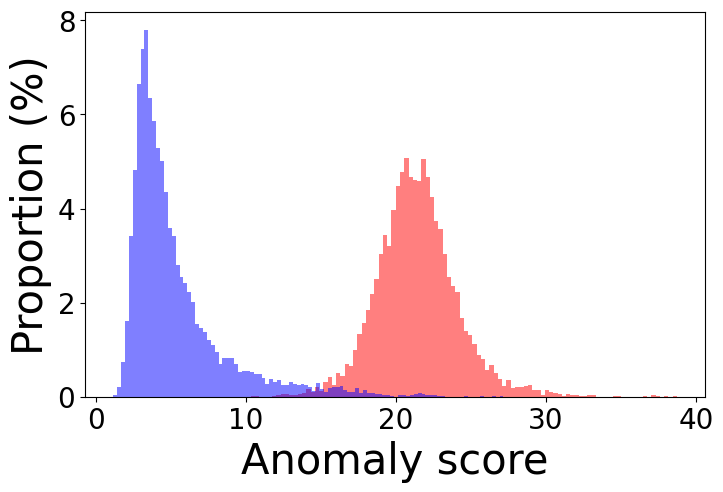}
    \caption{Imagenette}
    \label{fig:histogram_imagenette_data}
  \end{subfigure}%
  \begin{subfigure}[b]{0.2\columnwidth}
    \centering
    \includegraphics[width=\columnwidth]{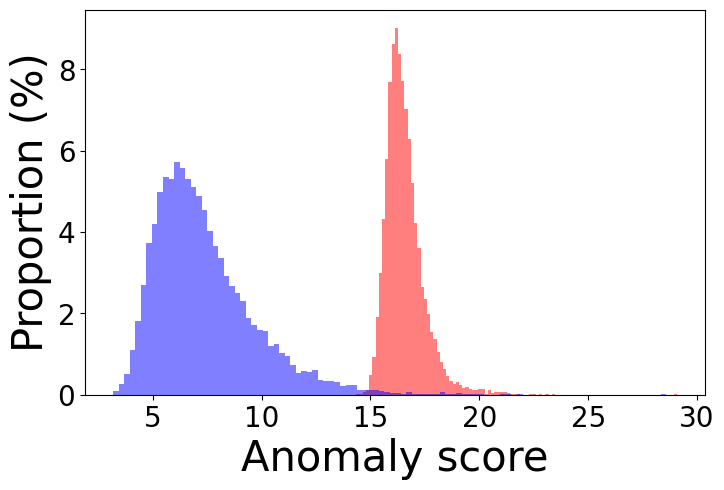}
    \caption{NUS-WIDE}
    \label{fig:histogram_data_nus}
  \end{subfigure}%
  \begin{subfigure}[b]{0.202\columnwidth}
    \centering
    \includegraphics[width=\columnwidth]{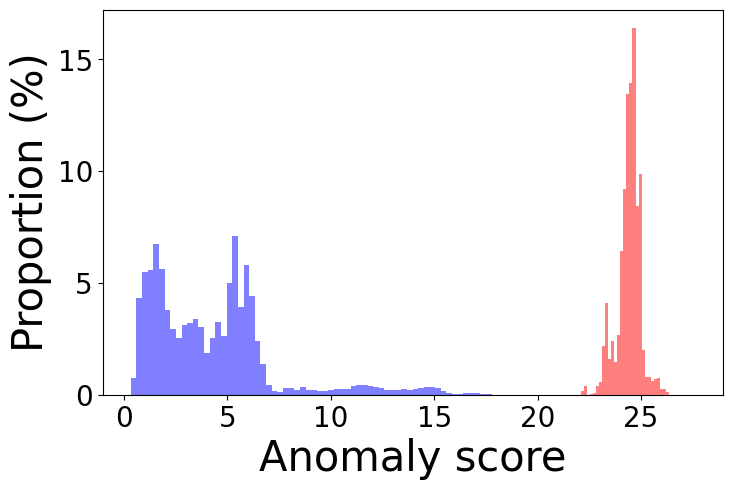}
    \caption{BM}
    \label{fig:histogram_bank_data}
  \end{subfigure}%
  \caption{Anomaly score distribution with BadVFL on five datasets.}
  \label{fig:score_bad}
\end{figure}
\begin{figure}[t!]
  \begin{subfigure}[b]{0.2\columnwidth}
    \centering
    \includegraphics[width=\columnwidth]{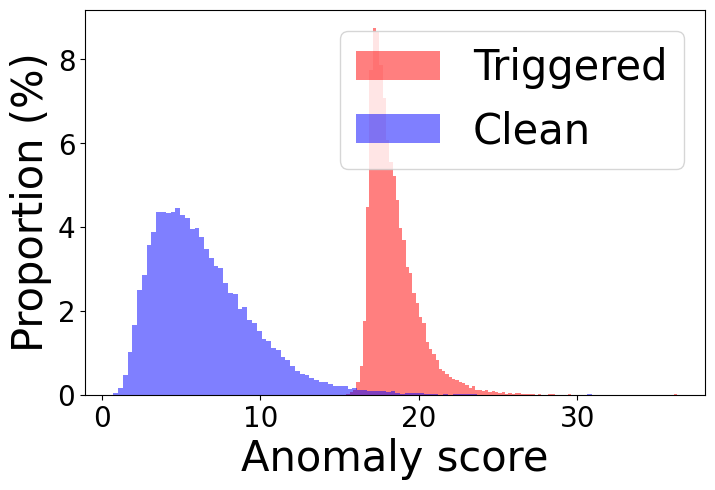} 
    \caption{CIFAR10}
    \label{fig:histogram_cifar_vil}
  \end{subfigure}%
  \begin{subfigure}[b]{0.2\columnwidth}
    \centering
    \includegraphics[width=\columnwidth]{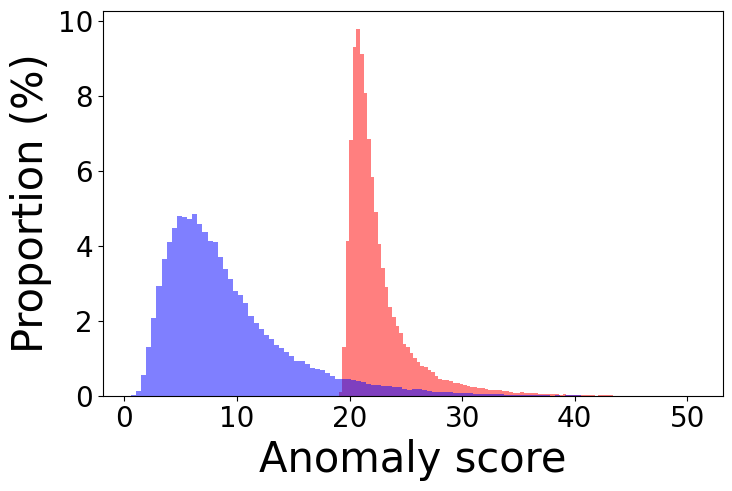}
    \caption{CINIC10}
    \label{fig:histogram_cinic_vil}
  \end{subfigure}%
  \begin{subfigure}[b]{0.2\columnwidth}
    \centering
    \includegraphics[width=\columnwidth]{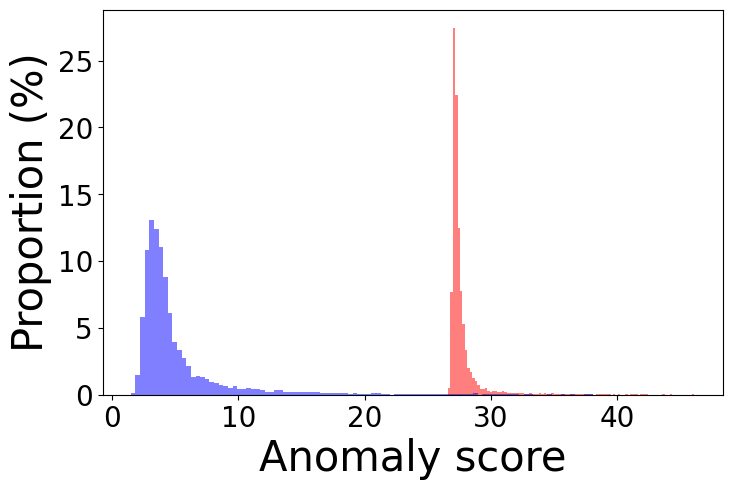}
    \caption{Imagenette}
    \label{fig:histogram_imagenette_vil}
  \end{subfigure}%
  \begin{subfigure}[b]{0.2\columnwidth}
    \centering
    \includegraphics[width=\columnwidth]{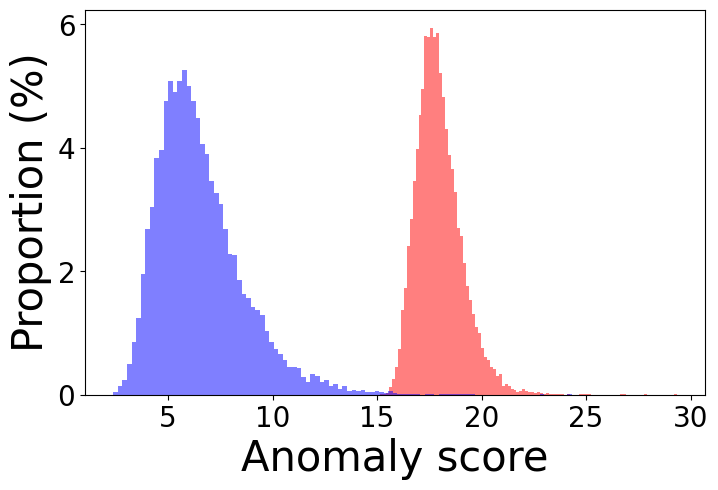}
    \caption{NUS-WIDE}
    \label{fig:histogram_vil_nus}
  \end{subfigure}%
  \begin{subfigure}[b]{0.20\columnwidth}
    \centering
    \includegraphics[width=\columnwidth]{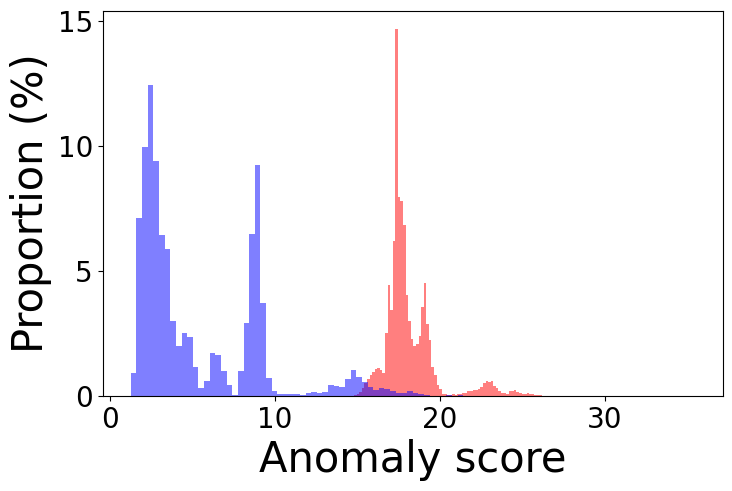}
    \caption{BM}
    \label{fig:histogram_bank_vil}
  \end{subfigure}%
  \caption{Anomaly score distribution with VILLAIN on five datasets.}
  \label{fig:score_vil}
  \vspace{-10pt}
\end{figure}

\subsection{Anomaly Score Distribution}
To demonstrate VFLIP's ability to identify backdoor-triggered embeddings based on the anomaly score, we visualize the anomaly score distribution of clean embeddings and backdoor-triggered embeddings. ~\autoref{fig:score_bad} and ~\autoref{fig:score_vil} present the distribution of the anomaly scores for each type of attack on five datasets. While there may be a slight overlap between the distributions, most backdoor-triggered embeddings are clearly separated from clean embeddings. 

\subsection{Ablation Study}\label{sec:ablation}
In this subsection, we assess how different factors affect the robustness of VFLIP. 
The default attack setting is the same as the main results with label-knowledgable attackers.

\begin{figure}[t]
    \centering
    \begin{subfigure}{\columnwidth}
        \centering
        \includegraphics[width=0.7\columnwidth]{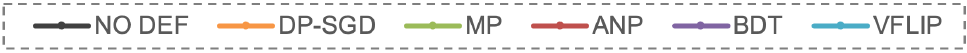}
    \end{subfigure}
    \begin{subfigure}{0.18\columnwidth}
        \centering
        \includegraphics[width=\columnwidth]{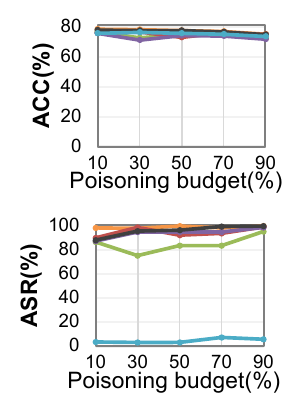}
        \caption{CIFAR10}
    \end{subfigure}
    \begin{subfigure}{0.18\columnwidth}
        \centering
        \includegraphics[width=\columnwidth]{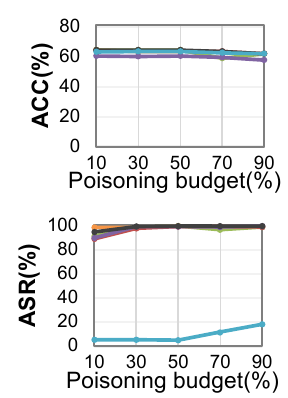}
        \caption{CINIC10}
    \end{subfigure}
    \begin{subfigure}{0.18\columnwidth}
        \centering
        \includegraphics[width=\columnwidth]{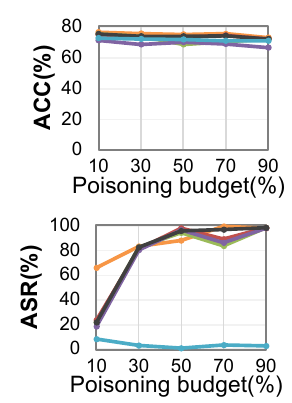}
        \caption{Imagenette}
    \end{subfigure}
    \begin{subfigure}{0.18\columnwidth}
        \centering
        \includegraphics[width=\columnwidth]{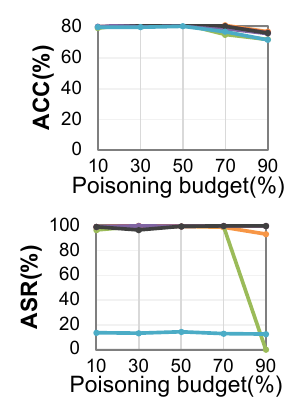}
        \caption{\hspace{-2pt}NUS-WIDE}
    \end{subfigure}
    \begin{subfigure}{0.18\columnwidth}
        \centering
        \includegraphics[width=\columnwidth]{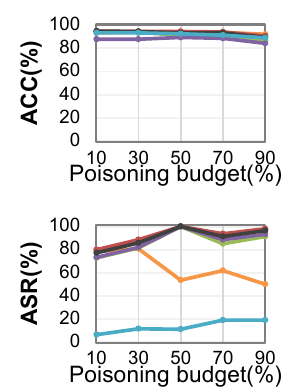}
        \caption{BM}
    \end{subfigure}
    \vspace{-3pt}
    \caption{Impact of poisoning budget with BadVFL on five datasets.}
    \label{fig:impact_poison_data}
    \vspace{-5pt}
\end{figure}
\begin{figure}[t]
    \centering
    \begin{subfigure}{\columnwidth}
        \centering
        \includegraphics[width=0.7\columnwidth]{figures/ablation/budget_legend.pdf}
    \end{subfigure}
    \begin{subfigure}{0.18\columnwidth}
        \centering
        \includegraphics[width=\columnwidth]{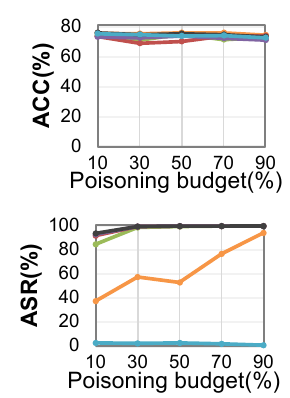}
        \caption{CIFAR10}
    \end{subfigure}
    \begin{subfigure}{0.18\columnwidth}
        \centering
        \includegraphics[width=\columnwidth]{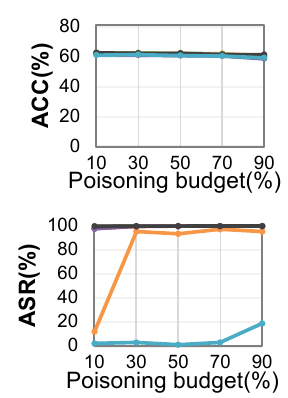}
        \caption{CINIC10}
    \end{subfigure}
    \begin{subfigure}{0.18\columnwidth}
        \centering
        \includegraphics[width=\columnwidth]{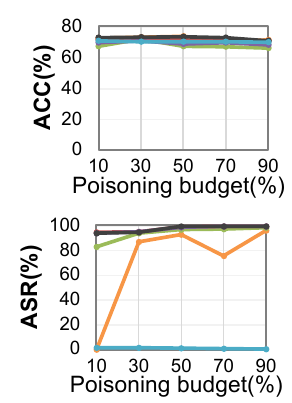}
        \caption{Imagenette}
    \end{subfigure}
    \begin{subfigure}{0.18\columnwidth}
        \centering
        \includegraphics[width=\columnwidth]{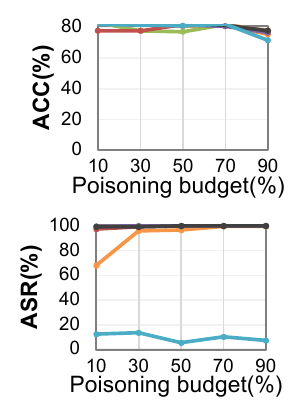}
        \caption{\hspace{-2pt}NUS-WIDE}
    \end{subfigure}
    \begin{subfigure}{0.18\columnwidth}
        \centering
        \includegraphics[width=\columnwidth]{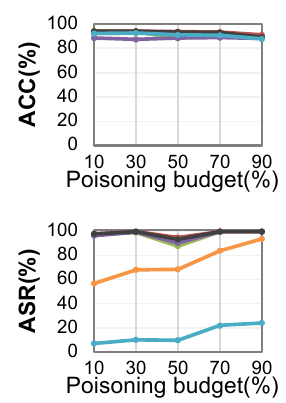}
        \caption{BM}
    \end{subfigure}
    \caption{Impact of poisoning budget with VILLAIN on five datasets.}
    \label{fig:impact_poison_vil}
    \vspace{-10pt}
\end{figure}
\begin{figure}[t]
    \centering
    \begin{subfigure}{\columnwidth}
        \centering
        \includegraphics[width=0.7\columnwidth]{figures/ablation/budget_legend.pdf}
    \end{subfigure}
    \begin{subfigure}{0.18\columnwidth}
        \centering
        \includegraphics[width=\columnwidth]{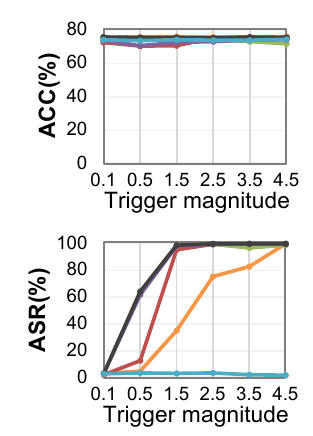}
        \caption{CIFAR10}
    \end{subfigure}
    \begin{subfigure}{0.18\columnwidth}
        \centering
        \includegraphics[width=\columnwidth]{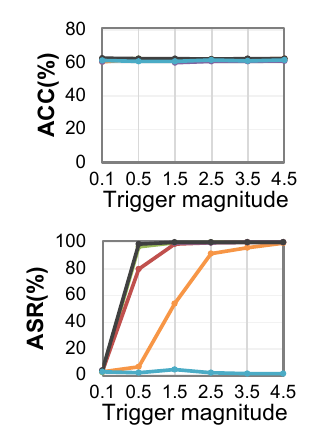}
        \caption{CINIC10}
    \end{subfigure}
    \begin{subfigure}{0.18\columnwidth}
        \centering
        \includegraphics[width=\columnwidth]{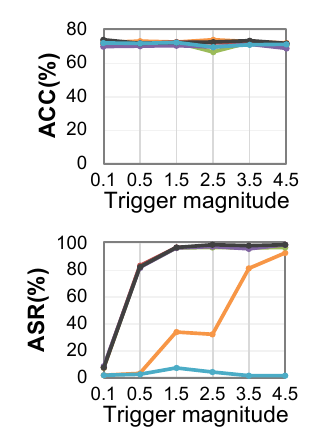}
        \caption{Imagenette}
    \end{subfigure}
    \begin{subfigure}{0.18\columnwidth}
        \centering
        \includegraphics[width=\columnwidth]{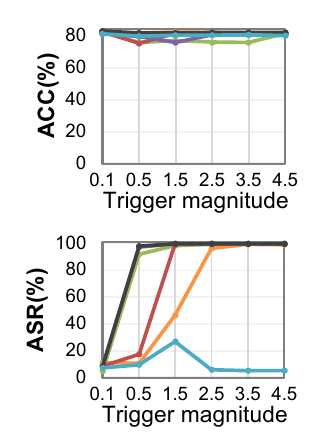}
        \caption{\hspace{-2pt}NUS-WIDE}
    \end{subfigure}
        \begin{subfigure}{0.18\columnwidth}
        \centering
        \includegraphics[width=\columnwidth]{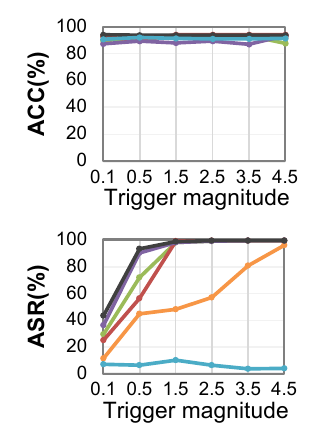}
        \caption{BM}
    \end{subfigure}
    \caption{Impact of trigger magnitude on five datasets.}
    \label{fig:impact_mag}
    \vspace{-10pt}
\end{figure}
\begin{figure}[t]
    \centering
    \begin{subfigure}{0.18\columnwidth}
        \centering
        \includegraphics[width=\columnwidth]{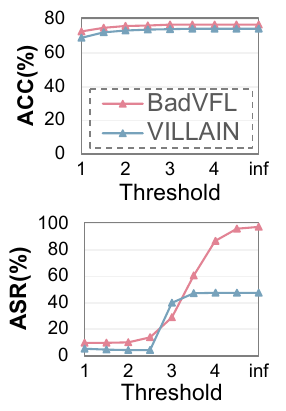}
        \caption{CIFAR10}
    \end{subfigure}
    \begin{subfigure}{0.18\columnwidth}
        \centering
        \includegraphics[width=\columnwidth]{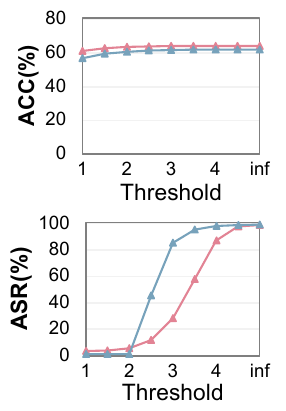}
        \caption{CINIC10}
    \end{subfigure}
    \begin{subfigure}{0.18\columnwidth}
        \centering
        \includegraphics[width=\columnwidth]{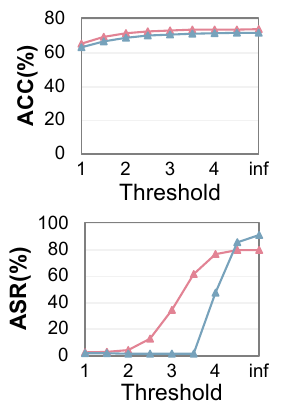}
        \caption{Imagenette}
    \end{subfigure}
    \begin{subfigure}{0.18\columnwidth}
        \centering
        \includegraphics[width=\columnwidth]{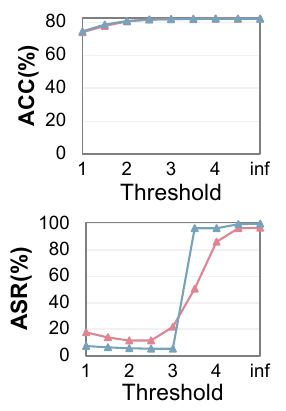}
        \caption{\hspace{-2pt}NUS-WIDE}
    \end{subfigure}
        \begin{subfigure}{0.18\columnwidth}
        \centering
        \includegraphics[width=\columnwidth]{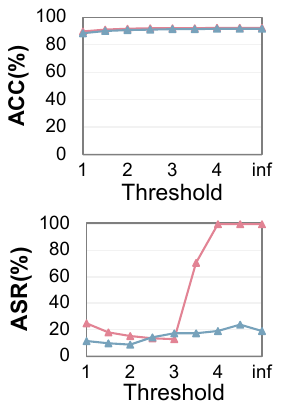}
        \caption{BM}
    \end{subfigure}
    \caption{Impact of anomaly score threshold $\rho$ for VFLIP on five datasets.}
    \label{fig:impact_thresh}
    \vspace{-20pt}
\end{figure}

\textbf{Impact of poisoning budget}.
In VFL, the backdoor attacker can easily adjust their poisoning budget. Therefore, considering an attacker who uses a large poisoning budget is essential.~\autoref{fig:impact_poison_data} and ~\autoref{fig:impact_poison_vil} presents the results of increasing the poisoning budget from 10\% to 90\%.
DP-SGD mitigates VILLAIN to some extent when the poisoning budget is small, but as the poisoning budget increases, the ASR significantly rises. Other defenses fail to mitigate both of the attacks in most cases. On the other hand, VFLIP shows stable defense performance across poisoning budgets. 

\textbf{Impact of trigger magnitude}.
Since there is no limit to the magnitude of the embeddings in VFL~\cite{liu2022copur}, VILLAIN can largely perturb the embeddings to increase the attack performance. Moreover, VILLAIN can attempt to evade identification mechanisms by sending a backdoor-triggered embedding that closely resembles a clean embedding using a small trigger magnitude. This requires the capability to defend against attacks with various trigger magnitudes.~\autoref{fig:impact_mag} presents experiments against VILLAIN with trigger magnitude ranging from 0.1 to 4.5. VFLIP outperforms other defense techniques in all datasets. This indicates that VFLIP can effectively mitigate attacks regardless of trigger magnitudes. Notably, VFLIP maintains the defensive performance even against the small-magnitude triggers (below scale 2). This is attributed to the purification phase which denoises the trigger by reconstructing all the embeddings. 


\textbf{Impact of bottom model architectures}.
The bottom model architecture can change depending on the attacker. Therefore, we evaluate various bottom model architectures. For image datasets, Resnet-20~\cite{he2016deep} and MobileNet~\cite{howard2017mobilenets} are evaluated. For NUS-WIDE and BM, 5-layer FCN and 3-layer FCN are evaluated. The results are provided in~\autoref{app:able_diff_arch}. In the experiments, VFLIP outperforms all other defenses. It indicates that VFLIP demonstrates the defense capability without relying on a specific bottom model architecture.

\textbf{Impact of the anomaly score threshold $\rho$ for VFLIP}. Most defense techniques face a trade-off between security and utility. In VFLIP, the threshold $\rho$ for the anomaly score introduces a trade-off between robustness and accuracy.~\autoref{fig:impact_thresh} illustrates the performance with respect to $\rho$. The results show that when $\rho$ is set to 2 or 2.5, there is a negligible decrease in ACC while ASR is still low. It indicates that there exists a proper trade-off point in VFLIP.

\textbf{Impact of the MAE training strategies}.
VFLIP uses two MAE training strategies. To evaluate each strategy individually, experiments are conducted by training MAE with only one strategy at a time. The results are provided in~\autoref{app:able_traing_strategy}. While using a single training strategy shows slightly better performance in some cases, using two training strategies shows more stable and better performance in most cases. 
\begin{figure}[t]
    \centering
    \begin{subfigure}[b]{0.18\columnwidth}
        \includegraphics[width=\columnwidth]{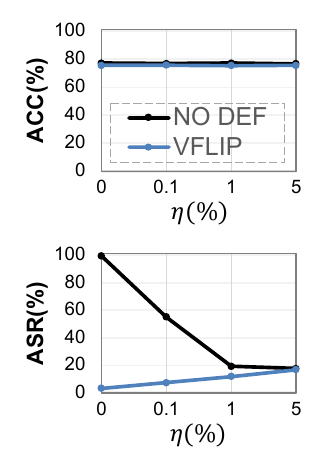}
        \caption{CIFAR10}
    \end{subfigure}
    \begin{subfigure}[b]{0.18\columnwidth}
        \includegraphics[width=\columnwidth]{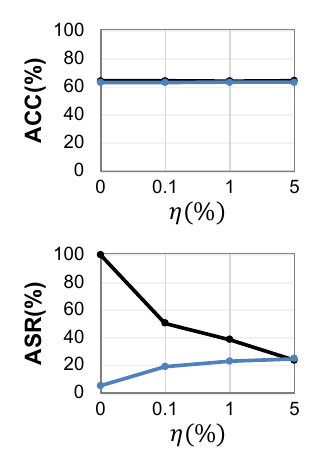}
        \caption{CINIC10}
    \end{subfigure}
    \begin{subfigure}[b]{0.18\columnwidth}
        \includegraphics[width=\columnwidth]{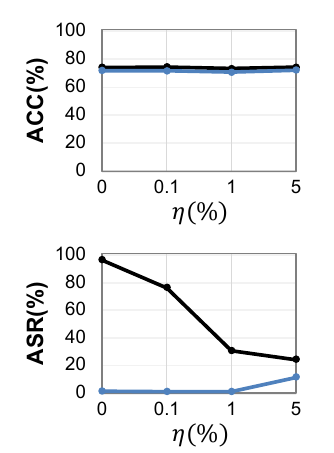}
        \caption{Imagenette}
    \end{subfigure}
    \begin{subfigure}[b]{0.18\columnwidth}
        \includegraphics[width=\columnwidth]{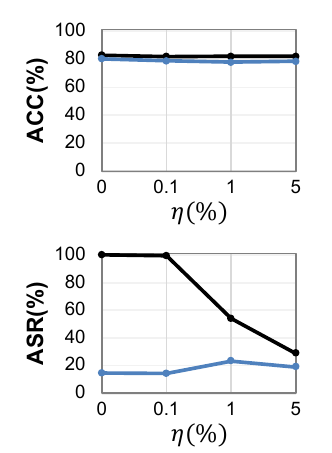}
        \caption{\hspace{-2pt}NUS-WIDE}
    \end{subfigure}
        \begin{subfigure}[b]{0.18\columnwidth}
        \includegraphics[width=\columnwidth]{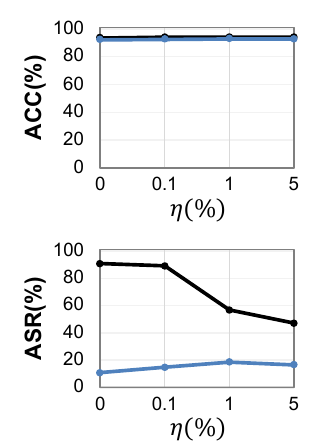}
        \caption{BM}
    \end{subfigure}
    \caption{Evaluation for the adaptive attacks with BadVFL against VFLIP.}
    \label{fig:adaptive_data}
\vspace{-10pt}
\end{figure}
\begin{figure}[t]
    \centering
    \begin{subfigure}[b]{0.18\columnwidth}
        \includegraphics[width=\columnwidth]{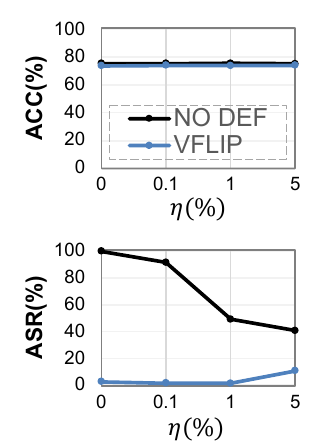}
        \caption{CIFAR10}
    \end{subfigure}
    \begin{subfigure}[b]{0.18\columnwidth}
        \includegraphics[width=\columnwidth]{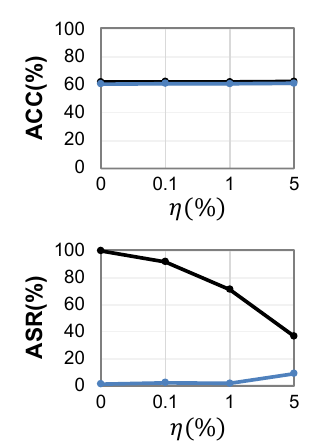}
        \caption{CINIC10}
    \end{subfigure}
    \begin{subfigure}[b]{0.18\columnwidth}
        \includegraphics[width=\columnwidth]{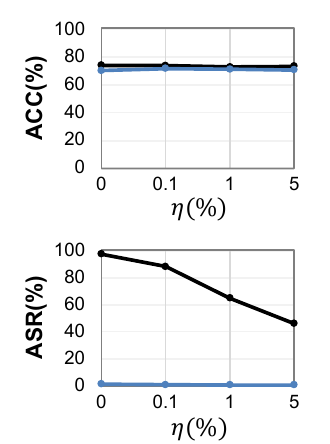}
        \caption{Imagenette}
    \end{subfigure}
    \begin{subfigure}[b]{0.18\columnwidth}
        \includegraphics[width=\columnwidth]{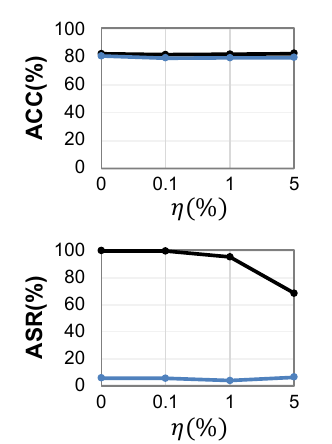}
        \caption{\hspace{-2pt}NUS-WIDE}
    \end{subfigure}
      \begin{subfigure}[b]{0.18\columnwidth}
        \includegraphics[width=\columnwidth]{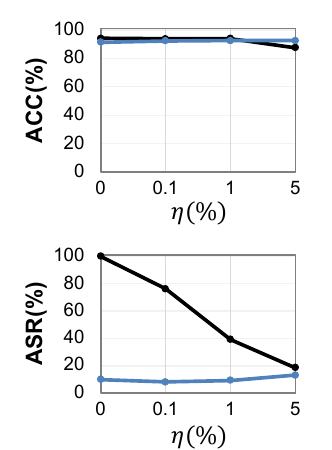}
        \caption{BM}
    \end{subfigure}
    \caption{Evaluation for the adaptive attacks with VILLAIN against VFLIP.}
    \label{fig:adaptive_vil}
    \vspace{-20pt}
\end{figure}

\section{Adaptive Attack}
To demonstrate the robustness of VFLIP against the attacker who knows the VFLIP mechanism, we design and evaluate an adaptive backdoor attack strategy. This strategy aims to poison $\mathcal{H}^{\operatorname{train}}$ so that the MAE learns relationships between the backdoor trigger and non-target embedding, reducing the anomaly scores. For this, at the last epoch, they insert backdoor triggers into samples of non-target labels with a certain probability. Thus, the abnormal cases are included in $\mathcal{H}^{\operatorname{train}}$. It makes VFLIP difficult to identify backdoor-triggered embeddings. Conversely, this weakens the connection between the backdoor trigger and the target label in the top model, reducing the attack success rate. To balance this trade-off, the attacker adjusts the probability of inserting triggers into non-target samples, denoted as $\eta$.

~\autoref{fig:adaptive_data} and~\autoref{fig:adaptive_vil} indicates that as $\eta$ increases, the ASR for VFLIP slightly increases. On the other hand, the ASR significantly declines when there is no defense. This implies that the connection between the backdoor trigger and the target label in the top model is substantially weakened before completely compromising the VFLIP's MAE. This demonstrates that to compromise the MAE, the attacker must sacrifice their ASR.

\section{Conclusion}
In this paper, we propose VFLIP, a novel backdoor defense for VFL.  VFLIP identifies the backdoor-triggered embedding and purifies their malicious influences. Additionally, we demonstrate that to compromise VFLIP's MAE, the attacker has to significantly sacrifice their ASR. Through extensive experiments, we demonstrate that VFLIP is robust and effective for defending against the backdoor attacks in the VFL. While this paper focuses on backdoor attacks in VFL, future research will need to explore these backdoor attacks in the broader context of VFL, considering additional complex problems such as non-IID issues.

\section{Acknowledgements}
This work was supported by the National Research Foundation of Korea(NRF) grant funded by the Korea government(MSIT) (RS-2023-00277326), Institute of Information \& communications Technology Planning \& Evaluation (IITP) grant funded by the Korea government(MSIT) [IITP-2023-RS-2023-00256081 (under the artificial intelligence semiconductor support program to nurture the best talents), No. 2022-0-00516 (Derivation of a Differential Privacy Concept Applicable to National Statistics Data While Guaranteeing the Utility of Statistical Analysis), RS-2021-II212068 (Artificial Intelligence Innovation Hub, EWU), and RS-2022-00155966 (Artificial Intelligence Convergence Innovation Human Resources Development, EWU)], the BK21 FOUR program of the Education and Research Program for Future ICT Pioneers, Seoul National University in 2024, Seoul R\&D Program(CY230117) through the Seoul Business Agency(SBA) funded by Seoul Metropolitan Government, 2024 AI Security Prototype Development Support Program funded by Ministry of Science and ICT and Korea Internet \& Security Agency, and azoo.ai.

\appendix
\vspace{-10pt}
\section{Appendix}

\subsection{VFL backdoor attacks}\label{Appendix:A}

\textbf{BadVFL}~\cite{xuan2023badvfl} is a data-level backdoor attack that consists of label inference and backdoor injection. 
For backdoor injection, BadVFL first replaces the target label local data with non-target label local data. Subsequently, the trigger is injected into the replaced data. BadVFL employs the pre-defined static trigger like the previous study~\cite{gu2017badnets}.\textbf{VILLAIN}~\cite{bai2023villain} is an embedding-level backdoor attack. Initially, VILLAIN conducts a label inference attack during the training stage. 
 For backdoor injection, VILLAIN utilizes trigger fabrication, backdoor augmentation, and learning rate adjustment. For trigger fabrication, VILLAIN chooses an additive trigger in the embedding level, rather than a replacement trigger. 
First, VILLAIN chooses $M$ dimensions with the highest standard deviation as the trigger area. Next, the trigger value is designed by using the average standard deviation of the selected dimension, denoted by $\sigma$. Ultimately, a repeating pattern of $\sigma$ serves as a backdoor trigger, represented as $\gamma \cdot [ \sigma, \sigma,- \sigma,- \sigma,\cdots,\sigma, \sigma,- \sigma,- \sigma]$ where $\gamma$ is a hyperparameter for the trigger magnitude. 
Next, the trigger undergoes two types of augmentation, during backdoor injection. One method involves randomly deleting parts of the trigger, and the other involves multiplying the trigger by a random value within the range of [$\underline{\lambda}$, $\bar{\lambda}$]. 
Moreover, before the backdoor injection, VILLAIN amplifies their local learning rate, causing the top model to be more dependent on the attacker's embeddings. Once the backdoor injection begins, the local learning rate is decreased to a smaller value.
The backdoor injection process is conducted after training for $E_{bkd}$ epochs. They empirically show that the existing backdoor defense for DNNs cannot mitigate VILLAIN. 

\vspace{-10pt}

\subsection{Attack Settings} \label{app:hyper_attack}
For the label inference attack~\cite{bai2023villain}, the label inference module~\cite{fu2022label} uses a batch size of 64 and a learning rate of 0.002. For swapping, the number of candidates in the minibatch is 3 $\times$ poisoning budget $\times$ batch size. If the gradient magnitude of the previous embedding is smaller than the average and the gradient magnitude of the swapped embedding is less than $10$ times the previous gradient magnitude, it is identified as a sample having the target label. Following previous studies~\cite{fu2022label,bai2023villain}, the attacker increases their local learning rate to enhance their malicious actions. For datasets other than Imagenette, the attacker's learning rate is multiplied by two, whereas it is increased by 1.2 times for Imagenette. The BadVFL trigger size is set to $5\times5$ for CIFAR10 and CINIC10, $40\times40$ for Imagenette, $60$ for NUS-WIDE, and  $8$ for BM. The VILLAIN trigger size is $75\%$ of the attacker's embedding dimension. The trigger magnitude $\gamma$ for VILLAIN is set to 3 for CIFAR10, CINIC10, NUS-WIDE, and BM, and 4 for Imagenette. We select the attacker as a participant holding features in the middle of the sample following the previous study~\cite{bai2023villain}.



\begin{table}[t!]
    \vspace{-5pt}
    \caption{Evaluation for various bottom model architectures.}
        \vspace{-10pt}
    \label{tbl:table_abl_arch}
     \begin{center}
\footnotesize
\renewcommand{\arraystretch}{0.9}
\resizebox{\textwidth}{!}{
\begin{tabular}{|T|T|C||WWWWWW||WWWWWU|}
     \hline
        \multirow{3}{*}[-0.8em]{Dataset} & \multirow{3}{*}[-0.8em]{Architecture} & \multirow{3}{*}[-0.8em]{Attack} & \multicolumn{12}{c|}{Defense} \\

        \cline{4-15}  
        
        {} & {} & {} & \multicolumn{6}{c||}{Accuracy (\%) $\uparrow$  \text{(Higher is better)}} & \multicolumn{6}{c|}{Attack Success Rate (\%) $\downarrow$ \text{(Lower is better)}} \\
        \cline{4-15} 
        
        {} & {} & {} & NO DEF & DP-SGD & \multirow{1}{*}[-0.7em]{MP}  & \multirow{1}{*}[-0.7em]{ANP} & \multirow{1}{*}[-0.7em]{BDT}  & \multirow{1}{*}[-0.7em]{VFLIP} & NO DEF & DP-SGD & \multirow{1}{*}[-0.7em]{MP}  & \multirow{1}{*}[-0.7em]{ANP} & \multirow{1}{*}[-0.7em]{BDT}  & \multirow{1}{*}[-0.7em]{VFLIP} \\
        \hline
        \hline

        \multirow{4}{*}[-0.1em]{CIFAR10} & \multirow{2}{*}{Resnet-20} &  BadVFL
        & \textbf{67.53}&66.38	&66.10	&65.86	&65.28	&65.03
        & 97.52	&95.33	&96.95	&92.36	&96.85	& \textbf{5.70}
        \\

        {} & {}  & VILLAIN 
        & 63.77	&62.00	&\textbf{63.78} &	62.97&	61.52&	62.00
        & 100.00	&81.93 &	100.00&	100.00	&100.00 &\textbf{3.74}
        \\
        \cline{2-15}

        {} & \multirow{2}{*}{MobileNet}  &  BadVFL 
        & 67.48&\textbf{68.60}&	65.36&	65.65&	62.07&	66.05
        & 99.26& 99.92&	88.46&	94.93&	98.15&	\textbf{7.88}
        \\

        {} & {}  & VILLAIN 
        & 64.39	&\textbf{65.10}&	62.98&	62.86&	61.84&	62.10
        & 100.00&	51.25&	100.00&	100.00&	100.00&	\textbf{2.22}
        \\

        \hline 
        \hline 
        \multirow{4}{*}[-0.1em]{CINIC10} & \multirow{2}{*}{Resnet-20}  &  BadVFL 
        & \textbf{54.70}&53.50&	50.18&	54.00&	52.42&	53.45
        & 99.37&98.59&	97.03&	99.68&	99.62&	\textbf{12.00}
        \\

        {} & {}  & VILLAIN 
        & \textbf{52.73}	&50.48&	49.33&	52.14&	49.74&	51.26
        & 100.00&	99.77&	99.97&	100.00&	100.00&	\textbf{2.28}
        \\
        \cline{2-15}

        {} & \multirow{2}{*}{MobileNet}  &  BadVFL 
        & 56.94&\textbf{57.20}&	56.24&	56.26&	54.64&	55.36
        & 99.92&99.96&	99.79&	99.88&	99.89&	\textbf{5.55}
        \\

        {} & {}  & VILLAIN 
        & \textbf{53.39}&52.23&	52.84&	52.90&	51.77&	51.67
        & 100.00&	100.00&	100.00&	100.00&	100.00&	\textbf{2.92}
        \\

        \hline 
        \hline 
        \multirow{4}{*}[-0.1em]{Imagenette} & \multirow{2}{*}{Resnet-20}  &  BadVFL 
        & \textbf{70.67}	&67.08&	66.66&	64.99&	68.27&	66.75
        & 87.19	&93.56&	80.67&	75.85&	86.48&	\textbf{24.36}
        \\

        {} & {}  & VILLAIN 
        & \textbf{67.82}&	63.32&	63.17&	65.74&	63.95&	63.37
        & 100.00&	99.83&	99.61&	99.91&	99.88&	\textbf{1.62}
        \\
        \cline{2-15}

        {} & \multirow{2}{*}{MobileNet}  &  BadVFL 
        &68.94	&\textbf{69.78}	&64.35	&69.23	&66.47	&65.44
        &96.13	&80.31	&68.73	&96.28	&95.62	&\textbf{12.47}
        \\

        {} & {}  & VILLAIN 
        & \textbf{69.36}	&67.37	&61.77	&62.50	&63.54	&66.67
        &99.74	&69.23	&99.70	&99.85	&99.82	&\textbf{2.90}
        \\

        \hline 
        \hline 
        \multirow{4}{*}[-0.1em]{NUS-WIDE} & \multirow{2}{*}{5-layer FCN}  &  BadVFL 
        & 80.03&	79.57&	75.08&	\textbf{80.49}&	76.86&	75.68
        & 98.69	&100.00&	68.73&	99.17&	98.81&	\textbf{24.28}
        \\

        {} & {}  & VILLAIN 
        & 81.32	&75.82&	\textbf{81.97}&	81.54&	81.12&	77.62
        & 100.00&	37.71&	100.00&	100.00&	100.00&	\textbf{5.84}
        \\
        \cline{2-15}

        {} & \multirow{2}{*}{3-layer FCN}  &  BadVFL 
        & \textbf{82.54}&81.91&	76.98&	82.49&	81.58&	80.65
        & 99.60&96.66&	99.88&	99.90&	99.94&	\textbf{9.47}
        \\

        {} & {}  & VILLAIN 
        & 82.75&	\textbf{83.32}&	76.66&	82.82&	81.29&	80.69
        & 100.00&	93.45&	88.19&	100.00&	99.98&	\textbf{5.07} \\

        \hline 
        \hline 
        \multirow{4}{*}[-0.1em]{BM} & \multirow{2}{*}{5-layer FCN}  &  BadVFL 
        & 93.64	&93.72&	89.03&	\textbf{94.00}&	86.93&	92.01&	89.80	&71.74&	71.15	&91.91&	84.82&	\textbf{14.96}
        \\

        {} & {}  & VILLAIN 
        & 93.72&	92.73&	91.71&	\textbf{94.05}	&87.78&	91.81&	99.90&	62.57	&99.66	&99.93&	99.55&	\textbf{14.80}
        \\
        \cline{2-15}

        {} & \multirow{2}{*}{3-layer FCN}  &  BadVFL 
        & 93.96&	93.36	&87.43&	\textbf{94.37}	&87.48	&91.34&	93.03	&64.50	&68.72&	94.75&	88.40&	\textbf{17.61}
        \\

        {} & {}  & VILLAIN 
        & 94.00&	91.68	&87.70&	\textbf{94.33}&	86.53	&89.28	&99.97&	85.33	&99.60&	100.00	&99.86&	\textbf{16.41} \\

        \hline
\end{tabular}
}
     \end{center}
      \vspace{-25pt}
\end{table}
\begin{table*}[t!]
    \vspace{-5pt}
 \caption{Evaluation for each training strategy.}
     \vspace{-10pt}
    \label{tbl:table_strategy}
   \begin{center}
    \renewcommand{\arraystretch}{0.9}
\resizebox{\textwidth}{!}{
\scriptsize
\begin{tabular}{|T|I|I||III||IIE|}
     \hline
        \multirow{3}{*}[-0.1em]{Dataset} & \multirow{3}{*}[-0.1em]{\makecell{Label\\ Knowledge}} & \multirow{3}{*}[-0.1em]{Attack} & \multicolumn{6}{c|}{Defense} \\

        \cline{4-9}  
        
        {} & {} & {} & \multicolumn{3}{c||}{Accuracy (\%) $\uparrow$  \text{(Higher is better)}} & \multicolumn{3}{c|}{Attack Success Rate (\%) $\downarrow$ \text{(Lower is better)}} 
        \\
        \cline{4-9} 
        
        {} & {} & {} & 1 to 1 & N-1 to 1  & VFLIP  & 1 to 1 & N-1 to 1  & VFLIP \\
 
        \hline
        \hline

        \multirow{4}{*}[-0.05em]{CIFAR10}  & \multirow{2}{*}{w/o}  & BadVFL 
        &\textbf{76.32}	&73.48	&75.62	
        &19.52	&\textbf{8.93}	&13.50
        \\

        {} & {}  & VILLAIN 
        &74.46	&71.09	&\textbf{75.33}
        & 3.94	&4.76	&\textbf{3.67}
        \\
        \cline{2-9}
        
        {} & \multirow{2}{*}{with}  & BadVFL 
        &\textbf{75.79}	&71.45	&75.56
        &3.75	&4.65	&\textbf{3.30}
        \\

        {} & {}  & VILLAIN 
        &73.33	&66.41	&\textbf{73.82}
        &4.42	&\textbf{1.99}	&2.78
        \\

        \hline 
        \hline
        \multirow{4}{*}[-0.05em]{CINIC10}  & \multirow{2}{*}{w/o}  & BadVFL 
        &\textbf{63.42}	&62.60	&63.18
        &17.74	&13.35	&\textbf{12.95}
        \\

        {} & {}  & VILLAIN 
       &62.18	&60.75	&\textbf{62.80}
        &2.94	&3.31	&\textbf{2.84}
        \\
        \cline{2-9}
        
        {} & \multirow{2}{*}{with}  & BadVFL 
        &63.04	&60.51	&\textbf{63.39}
        &6.32	&5.57	&\textbf{5.23}
        \\

        {} & {}  & VILLAIN 
        &\textbf{61.18}	&59.62	&60.74
        & 2.59	&2.13	&\textbf{1.17}
        \\

        \hline 
        \hline
        \multirow{4}{*}[-0.05em]{Imagenette} & \multirow{2}{*}{w/o}  & BadVFL 
        &71.74	&35.79	&\textbf{72.94}
        & 16.76	&42.49	&\textbf{14.00}
        \\

        {} & {}  & VILLAIN 
        &71.01	&25.05	&\textbf{71.32}
        &3.77	&26.42	&\textbf{2.68}
        \\
        \cline{2-9}
        
        {} & \multirow{2}{*}{with}  & BadVFL 
        &\textbf{72.73}	&41.28	&71.82
        &10.79	&22.91	&\textbf{1.31}
        \\

        {} & {}  & VILLAIN 
       &\textbf{71.01}	&28.73	&70.35
        &1.62	&0.00	&\textbf{1.28}
        \\

        \hline 
        \hline
        \multirow{4}{*}[-0.05em]{BM} & \multirow{2}{*}{w/o}  & BadVFL 
      &81.35	&79.46	&\textbf{81.65}
        & \textbf{9.39}	&29.83	&9.80
        \\

        {} & {}  & VILLAIN 
       &80.45	&80.23	&\textbf{81.51}
        &8.64	&10.64	&\textbf{6.38}
        \\
        \cline{2-9}
        
        {} & \multirow{2}{*}{with}  & BadVFL 
        &79.00	&78.54	&\textbf{80.31}
        &17.67	&21.44	&\textbf{14.54}
        \\

        {} & {}  & VILLAIN 
       &79.62	&79.80	&\textbf{80.83}
        &\textbf{4.82}	&8.28	&5.65
        \\

        \hline 
        \hline
        \multirow{4}{*}[-0.05em]{NUS-WIDE} & \multirow{2}{*}{w/o}  & BadVFL 
      &\textbf{93.47}	&49.97	&91.27
        & 15.97	& 0.00	&\textbf{14.71}
        \\

        {} & {}  & VILLAIN 
       &89.74 &49.97	&\textbf{91.91}
        &10.32	&8.28	&\textbf{7.31}
        \\
        \cline{2-9}
        
        {} & \multirow{2}{*}{with}  & BadVFL 
        &89.01	&50.00	&92.13
        &21.26	&100.00	&\textbf{10.78}
        \\

        {} & {}  & VILLAIN 
       &90.80	&50.00	&\textbf{91.30}
        &11.07	&100.00	&\textbf{9.82}
        \\

        \hline
\end{tabular}
}
    \end{center}
    \vspace{-30pt}
\end{table*}

\vspace{-10pt}
\subsection{Results for Label Inference Attacks}\label{app:able_label}
\vspace{-35pt}
\begin{table}[H]
    \caption{Accuracy of label inference attacks on five datasets.}
        \vspace{-10pt}
    \label{app:label_infer_acc}
    \begin{center}
    {
\resizebox{0.9\linewidth}{!}{
\def\arraystretch{1.2}
\begin{tabular}
{|c|CC|CC|CC|CC|CE|}
\hline

\multirow{4}{*}[0.5em]{\makecell{Label Inference\\ Attack \\  }}&
\multicolumn{10}{c|}{Label Inference Accuracy $\uparrow$  \text{(Higher is better)} }\\
\cline{2-11}
&  \multicolumn{2}{c|}{CIFAR10}
&  \multicolumn{2}{c|}{CINIC10}
&  \multicolumn{2}{c|}{Imagenette}
&  \multicolumn{2}{c|}{NUS-WIDE}
&  \multicolumn{2}{c|}{BM}
\\

\cline{2-11}

& DP-SGD&		Others&
DP-SGD&		Others&
DP-SGD&		Others&
DP-SGD&		Others&
DP-SGD&		Others  \\
\hline



\cite{bai2023villain}&
88.16&	89.27&
94.59&	93.15&
98.97&	95.20&
94.16&	94.85&
83.63 & 89.14
\\

\hline

\end{tabular}
}
}

    \end{center}
    \vspace{-30pt}
\end{table}




\subsection{Impact of Bottom Model Architecture}\label{app:able_diff_arch}
\vspace{-5pt}
~\autoref{tbl:table_abl_arch} presents the results for various bottom architectures. 
\vspace{-15pt}

\subsection{Impact of the MAE Training Strategies}\label{app:able_traing_strategy}
    \vspace{-5pt}
~\autoref{tbl:table_strategy} presents the results for each training strategy in VFLIP. 
\vspace{-15pt}

\bibliographystyle{splncs04.bst}
\bibliography{mybibliography.bib}
\end{document}